\documentclass{article} 
\usepackage{iclr2023_conference,times}

\usepackage{amsmath,amsfonts,bm}

\def\eqref#1{equation~\ref{#1}}

\def\1{\bm{1}}

\DeclareMathAlphabet{\mathsfit}{\encodingdefault}{\sfdefault}{m}{sl}
\SetMathAlphabet{\mathsfit}{bold}{\encodingdefault}{\sfdefault}{bx}{n}

\usepackage{hyperref}
\usepackage{url}

\hypersetup{
  colorlinks   = true, 
  urlcolor     = red, 
  linkcolor    = [rgb]{0,0,0.5}, 
  citecolor   = [rgb]{0,0,0.5} 
}

\usepackage[utf8]{inputenc} 
\usepackage[T1]{fontenc}    
\usepackage{hyperref}       
\usepackage{url}            
\usepackage{booktabs}       
\usepackage{amsfonts}       
\usepackage{nicefrac}       
\usepackage{microtype}      
\usepackage{xcolor}         

\usepackage{booktabs}
\usepackage{multirow}
\usepackage{lipsum}
\usepackage{wrapfig}
\usepackage{here}    
\usepackage{amsmath}
\usepackage{amssymb}
\usepackage{microtype}
\usepackage{graphicx}
\usepackage{booktabs} 
\usepackage{bm}
\usepackage{amsthm}
\newtheorem{theorem}{Theorem}[section]
\newtheorem{lemma}[theorem]{Lemma}
\newtheorem{proposition}[theorem]{Proposition}

\newtheorem{assumption}[theorem]{Assumption}

\usepackage{color}
\usepackage{lipsum}

\usepackage{algorithm}

\usepackage{algorithmic}

\usepackage{caption}
\usepackage{subcaption}

\newcommand{\com}[1]{\textcolor{black}{#1}}

\usepackage{wrapfig}

\newcommand*{\QEDA}{\hfill\ensuremath{\blacksquare}}%

\usepackage{listings}
\usepackage{xcolor}
\definecolor{codegreen}{rgb}{0,0.6,0}
\definecolor{codegray}{rgb}{0.5,0.5,0.5}
\definecolor{codepurple}{rgb}{0.58,0,0.82}
\definecolor{backcolour}{rgb}{0.95,0.95,0.92}

\lstdefinestyle{mystyle}{
  backgroundcolor=\color{backcolour}, commentstyle=\color{codegreen},
  keywordstyle=\color{magenta},
  numberstyle=\tiny\color{codegray},
  stringstyle=\color{codepurple},
  basicstyle=\ttfamily\footnotesize,
  breakatwhitespace=false,         
  breaklines=true,                 
  captionpos=b,                    
  keepspaces=true,                 
  numbers=left,                    
  numbersep=5pt,                  
  showspaces=false,                
  showstringspaces=false,
  showtabs=false,                  
  tabsize=2
}
\title{Understanding Gradient Regularization in Deep Learning: Efficient Finite-Difference Computation and  Implicit Bias}

\author{%
 Ryo Karakida
 \thanks{ E-mail: \texttt{karakida.ryo@aist.go.jp}}\\
    AIST, Japan\\
    \And
  Tomoumi Takase\\
    AIST, Japan\\
        \And
  Tomohiro Hayase\\
    Cluster, Japan\\
    \And
    Kazuki Osawa\\
    ETH Zurich, Switzerland\\    
}

\iclrfinalcopy

\begin{document}
\maketitle

\begin{abstract}
Gradient regularization (GR) is a method that penalizes the gradient norm of the training loss during training. While some studies have reported that GR can improve generalization performance, little attention has been paid to it from the algorithmic perspective, that is, the algorithms of GR that efficiently improve the performance. In this study, we first reveal that a specific finite-difference computation, composed of both gradient ascent and descent steps, reduces the computational cost of GR. Next, we show that the finite-difference computation also works better in the sense of generalization performance. We theoretically analyze a solvable model, a diagonal linear network, and clarify that GR has a desirable implicit bias to so-called rich regime and finite-difference computation strengthens this bias. Furthermore, finite-difference GR is closely related to some other algorithms based on iterative ascent and descent steps for exploring flat minima. In particular, we reveal that the flooding method can perform finite-difference GR in an implicit way. Thus, this work broadens our understanding of GR for both practice and theory.
\end{abstract}

\section{Introduction}

Explicit or implicit regularization is a key component for achieving better performance in deep learning. 
For instance, adding some regularization on the local sharpness of the loss surface is one common approach to enable the trained model to achieve better performance \citep{hochreiter1997flat,foret2020sharpness,jastrzebski2021}. In the related literature, some recent studies have empirically reported that gradient regularization (GR), i.e., adding  penalty of the gradient norm to the original loss, makes the training dynamics reach flat minima and leads to better generalization performance \citep{barrett2020implicit,smith2021origin,zhao22i}. Using only the information of the first-order gradient seems a simple and computationally friendly idea.
 Because the first-order gradient is used to optimize the original loss, using its norm is seemingly easier to use than other sharpness penalties based on second-order information such as the Hessian and Fisher information \citep{hochreiter1997flat,jastrzebski2021}. 

Despite its simplicity, our understanding of GR has been limited so far in the following points. First, we need to consider the fact that GR must compute {\it the gradient of the gradient} with respect to the parameter. This type of computation has been investigated in a slightly different context: input-Jacobian regularization, that is, penalizing the gradient with respect to the input dimension to increase robustness against input noise  \citep{drucker1992improving,hoffman2019robust}. Some studies proposed the use of double backpropagation (DB) as an efficient algorithm for computing the gradient of the gradient for input-Jacobian regularization, whereas others proposed the use of finite-difference computation \citep{peebles2020hessian,finlay2020scaleable}. 
It remains unclear which algorithm is more efficient in the case of GR. 
Second, theoretical understanding of GR has been limited. Although empirical studies have confirmed that the GR causes the gradient dynamics to eventually converge to better minima with higher performance, the previous work provides no concrete theoretical evaluation for this result. Third, it also remains unclear whether the GR has any potential connection to other regularization methods. Because the finite difference is composed of both gradient ascent and descent steps by definition, we are reminded of some learning algorithms for exploring flat minima such as sharpness-aware minimization (SAM) \citep{foret2020sharpness} and the flooding method \citep{ishida2020we}, which are also composed of ascent and descent steps. Clarifying these points would help to deepen our understanding of efficient regularization methods for deep learning.

In this work, we reveal that a finite-difference computation of GR works efficiently. This approach has a lower computational cost, and surprisingly achieves better generalization performance. 
We present three main contributions to deepen our understanding of GR: 
\begin{itemize}
\item We demonstrate some advantages to using the finite-difference computation. We give a brief estimation of the computational costs of finite difference and DB in a deep neural network, and show that the finite difference is more efficient than DB (Section \ref{Sec3_2fin}). 

\item 
We find that a so-called forward finite difference leads to better generalization than a backward one and DB  (Section \ref{Sec3_3}). 
Learning with forward finite-difference GR requires two gradients of the loss function,  gradient ascent and descent. We reveal that a relatively large positive ascent step improves the generalization.  
In addition, we give a theoretical analysis of the performance improvement obtained by finite-difference GR. 
We analyze the selection of global minima in a diagonal linear network (DLN), which is a theoretically solvable model. We prove that GR has an implicit bias for selecting desirable solutions in the so-called rich regime \citep{woodworth2020kernel} which would potentially lead to better generalization (Section \ref{Sec4_2f}). This implicit bias is strengthened when we use  forward finite-difference GR with an increasing ascent step size. In contrast, it is weakened for a backward finite difference, i.e., a negative ascent step. 

\item  Finite-difference GR is also closely related to other learning methods composed of both gradient ascent and descent.  
In particular, we reveal that the flooding method performs finite-difference GR in an implicit way (Section \ref{Sec5_2}). 
\end{itemize}
Thus, this work gives a comprehensive perspective on GR for both practical and theoretical understanding.

\section{Preliminaries}
\label{sec3}
\subsection{Gradient Regularization}

We consider GR \citep{barrett2020implicit,smith2021origin}, wherein the squared L2 norm of the gradient is explicitly added to the original loss $\mathcal{L}(\theta)$ as follows:
\begin{equation}
\tilde{\mathcal{L}}(\theta) = \mathcal{L}(\theta) + \frac{\gamma}{2} R(\theta), \ \  R(\theta) = \|\nabla \mathcal{L}(\theta) \|^2, \label{eq20:0324}
\end{equation}
where $\| \cdot\|$ denotes the Euclidean norm and $\gamma>0$ is a constant regularization coefficient. 
We abbreviate the derivative with respect to the parameters $\nabla_\theta$ by $\nabla$. Its gradient descent is given by  
\begin{equation}
\theta_{t+1} =\theta_t - \eta \nabla \tilde{\mathcal{L}}(\theta_t) \ \ \label{eq2}
\end{equation}
for time step $t=0,1,...$ and learning rate $\eta>0$. While previous studies have reported that explicitly adding a GR term empirically improves generalization performance, its  algorithms and implementations have not been discussed in much detail.

\subsection{Algorithms}
\noindent

To optimize the loss function with GR (\ref{eq20:0324}) using a gradient method, we need to compute the gradient of the gradient, i.e., $\nabla R(\theta)$.  
As is well studied in input-Jacobian regularization \citep{drucker1992improving,hoffman2019robust,finlay2020scaleable}, there are two main approaches to computing the gradient of the gradient.

\noindent
{\bf Finite difference: }
The finite-difference method approximates a derivative by a finite step. In the case of GR, we have  
$\nabla R(\theta_t)/2 =  (\nabla \mathcal{L}(\theta') -\nabla \mathcal{L}(\theta_t))/\varepsilon + \mathcal{O}(\varepsilon)$ 
with $\theta'= \theta_t + \varepsilon \nabla \mathcal{L}(\theta_t)$ for a constant $\varepsilon>0$. 
The final term is expressed in Landau notation and is neglected in the computation. We update the GR term by 
\begin{equation}
\Delta R_F(\varepsilon) =  \frac{\nabla \mathcal{L}(\theta_t + \varepsilon \nabla \mathcal{L}(\theta_t)) -\nabla \mathcal{L}(\theta_t)}{\varepsilon} \quad (\text{F-GR}).  \label{eq18:0428}
\end{equation}
We refer to this gradient as {\it Forward finite-difference GR (F-GR)}. Because the gradient $\nabla \mathcal{L}(\theta_t)$ is computed for the original loss, the finite difference (\ref{eq18:0428}) requires only one additional gradient computation $\nabla \mathcal{L}(\theta')$. The order of the computation time is only double that of the usual gradient descent. 
The finite-difference method also has a backward computation: 
\begin{equation}
\Delta R_B(\varepsilon) =  \frac{\nabla \mathcal{L}(\theta_t)- \nabla \mathcal{L}(\theta_t - \varepsilon \nabla \mathcal{L}(\theta_t)) }{\varepsilon} \quad (\text{B-GR}).  \label{eq18:0428B}
\end{equation}
If we allow a negative step size, $\Delta R_B$ corresponds to $\Delta R_F$ through $\Delta R_B (\varepsilon)= \Delta R_F (-\varepsilon)$. 

\begin{figure*}
\vspace{5pt}
  \begin{center}
\includegraphics[width=1\textwidth]{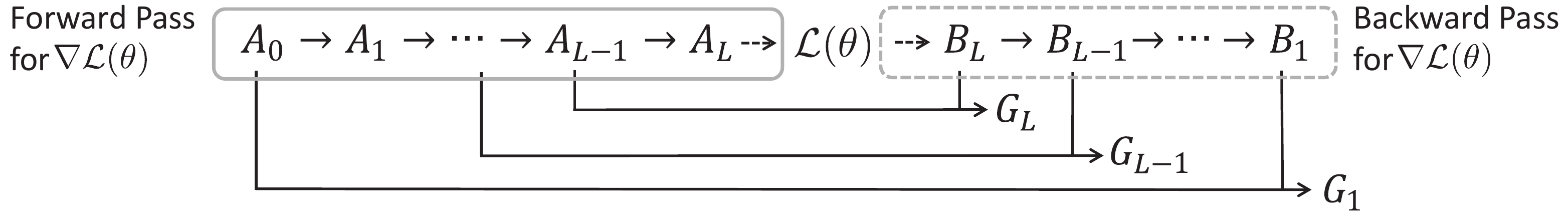}
     \end{center}
\caption{Computational graph of DB. Each node with an incoming solid arrow requires one matrix multiplication for the forward pass.}
\label{fig1}
\end{figure*}

\begin{figure*}
    \centering
    \includegraphics[width=0.9\textwidth]{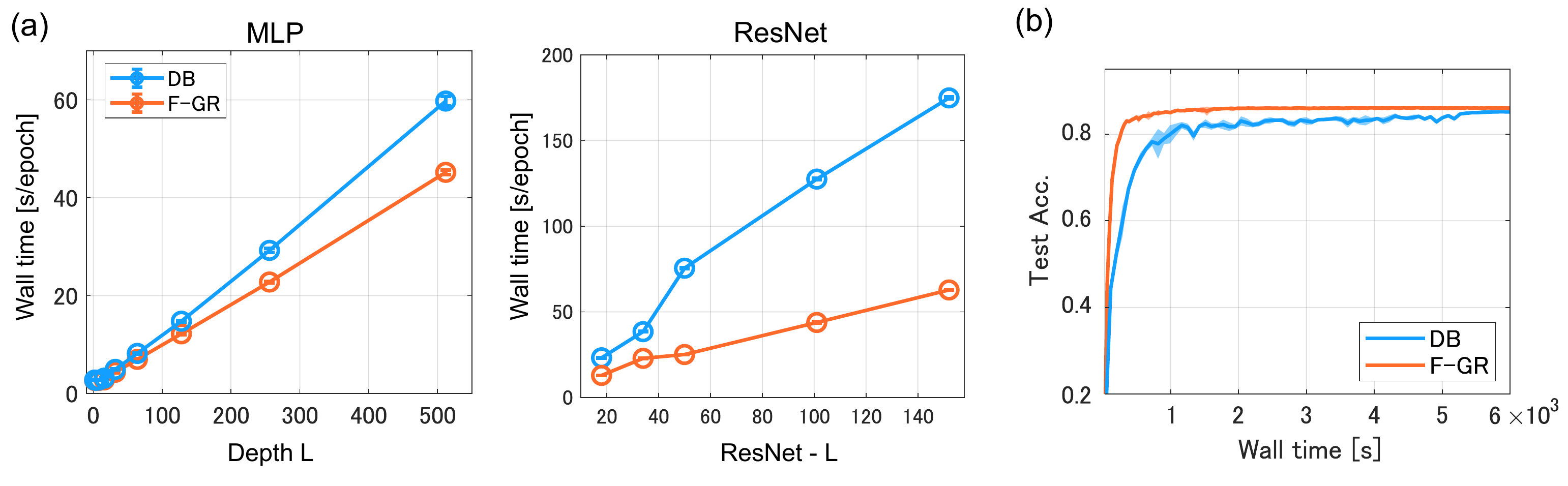}
    \caption{Finite-difference computation is more efficient than DB computation in wall time. (a) Wall time required for learning with GR in one epoch. For the ResNet, we used  ResNet-$\{18,34,50,101,152\}$. 
    (b)Training dynamics in ResNet-18 on CIFAR-10. Learning with F-GR is much faster in wall time.}
    \label{fig2}
\end{figure*}

{\bf Double Backpropagation: } The other approach is to apply the automatic differentiation directly to the GR term, i.e., $\nabla R$. For example, its PyTorch implementation is quite straightforward, as shown in Section \ref{pcode} of the Appendices. This approach is referred to as DB, which was originally developed for input-Jacobian regularization \citep{drucker1992improving}. 
We explain more details on the DB computation and its computational graph in Section \ref{Sec3_2fin}. 
DB, in effect, corresponds to computing the original gradient $\nabla R(\theta)$ given by the following Hessian-vector product:
\begin{equation}
 \Delta R_{DB} = H(\theta_t) \nabla \mathcal{L}(\theta_t), \label{eq5:0918}
\end{equation}
where $H(\theta)=\nabla \nabla \mathcal{L}(\theta)$.

Note that for a sufficiently small $\varepsilon$,  finite-difference GRs yield the same original gradient $\nabla R(\theta)$ if we can neglect any numerical instability caused by the limit. 
The finite-difference method has been used in the literature for the optimization of neural networks, especially for Hessian-based  techniques \citep{bishop2006pattern,peebles2020hessian}.  When we need a more precise $\nabla R$, we can use a higher-order approximation, e.g., the centered 
finite difference, but this requires additional gradient computations, and hence we focus on the first-order finite difference.  

\subsection{Related Work}
\citet{barrett2020implicit} and \citet{smith2021origin} investigated explicit and implicit GR in deep learning. They found that the discrete-time update of the usual gradient descent implicitly regularizes the gradient norm when its dynamics are mapped to the continual-time counterpart. This is referred to as implicit GR.  
They also investigated explicit GR, i.e., adding a GR term explicitly to the original loss, and reported that it improved generalization performance even further. \citet{jia2020information} also empirically confirmed that the explicit GR gave the improvement of generalization.
 \citet{barrett2020implicit} characterized GR as the slope of the loss surface and showed that a low GR (gentle slope) prefers flat regions of the surface. 
Recently, \citet{zhao22i} independently proposed a similar but different gradient norm regularization, that is, explicitly adding a non-squared L2 norm of the gradient to the original loss.  
They used a forward finite-difference computation, but its superiority to other computation methods remains unconfirmed.

The implementation of GR has not been discussed in much detail in the literature. In general, to compute the gradient of the gradient, 
there are two well-known computational methods: DB and finite difference.
Some previous studies applied DB to the regularization of an information matrix \citep{jastrzebski2021} and input-Jacobian regularization,  i.e., adding the L2 norm of the derivative with respect to the input dimension \citep{drucker1992improving,hoffman2019robust}. Others have used the finite-difference computation for  Hessian regularization \citep{peebles2020hessian} and 
input-Jacobian regularization \citep{finlay2020scaleable}.
Here, we apply the finite-difference computation to GR and present some evidence that the finite-difference computation outperforms DB computation with respect to computational costs and generalization performance.

In Section 4, we give a theoretical analysis of learning with GR in {\it diagonal linear networks} (DLNs) \citep{woodworth2020kernel}.
The characteristic property of this solvable model 
is that we can evaluate the implicit bias of learning algorithms \citep{nacson2022implicit,pesme2021implicit}. Our analysis includes the analysis of SAM in DLN as a special case \citep{andriushchenko2022towards}. In contrast to previous work, we evaluate some novel terms caused by the finite ascent step size, and this enables us to show that forward finite-difference GR selects global minima in the so-called rich regime.

\section{Computational Aspect}
\label{Sec3_2fin}
We clarify the computational efficiencies of each algorithm of GR in deep networks. 
First, we give a rough estimation of the computational cost by counting  the number of matrix multiplication required to compute $\nabla \tilde{\mathcal{L}}$. Consider an $L$-layer fully connected neural network with a linear output layer: $A_{l}=\phi (U_{l})$, $U_{l}= W_{l} A_{l-1}$ for $l=1,...,L$. Note that $A_l$ denotes a batch of activation and $W_{l} A_{l-1}$ requires a matrix multiplication.
We denote the element-wise activation function as $\phi(\cdot)$ and weight matrix as $W_{l}$. For simplicity, we neglect the bias terms. 
The number of matrix multiplications required to compute $\nabla \tilde{\mathcal{L}}$ is given by
\begin{equation}
N_{mul} \sim 6L \ (\text{for F-GR}), \ \    9L \ (\text{for DB}), \label{eq7:0926}
\end{equation}
where $\sim$ hides an uninteresting constant shift independent of the depth. 
 One can evaluate $N_{mul}$ straightforwardly from the computational graph (Figure \ref{fig1}), originally developed for the DB computation of input-Jacobian regularization \citep{drucker1992improving}. In brief, the original gradient $\nabla \mathcal{L}$, that is, 
 the backpropagation on the forward pass $\{A_0 \rightarrow A_1 \rightarrow \cdots \rightarrow A_L  \}$, requires $3L$ matrix multiplications: $L$ for the forward pass, $L$ for backward pass $B_{l} = \phi'(U_{l}) \circ  (W_{l+1}^\top B_{l+1})$, and $L$ for gradient $G_l:=\partial \mathcal{L}/\partial W_l = B_l A_{l-1}^\top$. 
 Because F-GR is composed of both gradient ascent and descent steps, we eventually need $6L$. In contrast, for learning using the DB of GR, we need $3L$ for $\nabla \mathcal{L}$ and additional $6L$ for the GR term. The GR term requires a forward pass of composed of $A_l$, $B_l$, and $G_l$ obtained in the gradient computation of $\nabla \mathcal{L}$. Note that the upper part $\{A_0 \rightarrow A_1 \rightarrow \cdots \rightarrow B_L  \rightarrow \cdots \rightarrow B_1 \}$ is well known as the DB of input-Jacobian regularization. As pointed out in \citet{drucker1992improving}, the computation of $\nabla B_{1}$ is equivalent to treating the upper part of the graph as the forward pass and applying backpropagation. It requires $2L$ multiplications.  In our GR case, we have additional $L$ multiplications due to $G_l$. Because the backward pass doubles the number of required multiplications, we eventually need $2 \times (2L+L)=6L$ multiplication. 
Further details are given in Section \ref{S_matmul}.

The results of numerical experiments shown in Figure \ref{fig2} confirm the superiority of finite-difference GR in typical experimental settings. We trained deep neural networks using an NVIDIA A100 GPU for this experiment. All experiments were implemented by PyTorch.  We summarize the pseudo code  and implementation of GR in Section \ref{pcode} and present the detailed settings of all experiments in Section \ref{Sec_C2}.  
Figure \ref{fig2}(a) shows the wall time required for one epoch of training with stochastic gradient descent (SGD) and the objective function (\ref{eq20:0324}). We trained various multi-layer perceptrons (MLPs) and residual neural networks (ResNets) with different depths. 
The wall time increased almost linearly as the depth increased. The slope of the line is different for F-GR and DB, and F-GR was faster. This observation is consistent with the number of multiplications (\ref{eq7:0926}). In particular, in ResNet, one of the most typical deep neural networks, learning with finite-difference GR was more than twice as fast as learning with DB. Figure \ref{fig2}(b) confirms that F-GR has fast convergence in ResNet-18 on CIFAR-10. In Figure \ref{figS1}, we also show the convergence measured by the training loss and time steps. All of them showed better convergence for the finite difference. 
 
Note that the finite difference is also better to use from the perspective of memory efficiency. This is because DB requires all of the $\{A_l,B_l,G_l\}$ to be retained for the forward pass, which occupies more memory.  
It is also noteworthy that in general, it is difficult for theory to completely predict the realistic computational time required because it could heavily depend on the hardware and the implementation framework and does not necessarily correlate well with 
the number of floating-point operations (FLOPs) \citep{dehghani2021efficiency}. 
Our result suggests that at least the number of matrix multiplication explains well the superiority of the finite-difference approach in typical settings.

\section{Implicit Bias of GR}
\label{sec4_fin}

In this section, we show that the superiority of finite-difference computation over DB also appears in the eventual performance of trained models. First, we show the empirical results that F-GR with a relatively large step size 
achieves better generalization performance. Next, we confirm this superiority in a solvable 
 network model that is non-linear with respect to parameters.

\subsection{Empirical Observation of Trained Models}
\label{Sec3_3}

\begin{figure}[t]
\centering
    \includegraphics[width=1\textwidth]{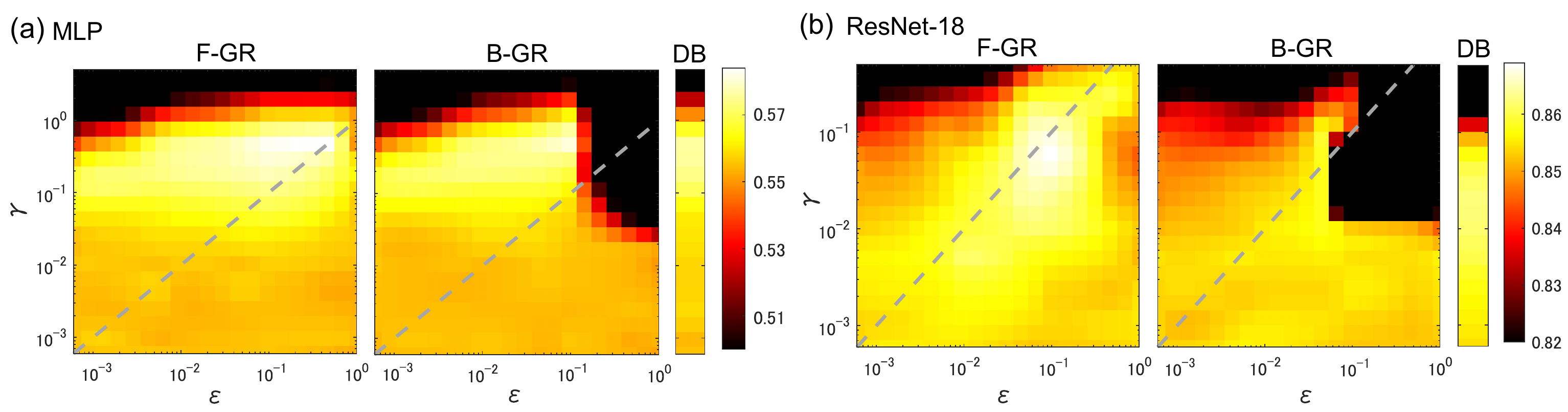}
\caption{Grid search on learning with different GR algorithms shows the superiority of F-GR and that a relatively large $\varepsilon$ achieves a high test accuracy. 
The color bar shows the average test accuracy over 5 trials. Gray dashed lines indicate $\gamma = \varepsilon$.}
\vspace{-5pt}
\label{fig4}
\end{figure}

\begin{figure}[t]
\centering
    \includegraphics[width=0.8\textwidth]{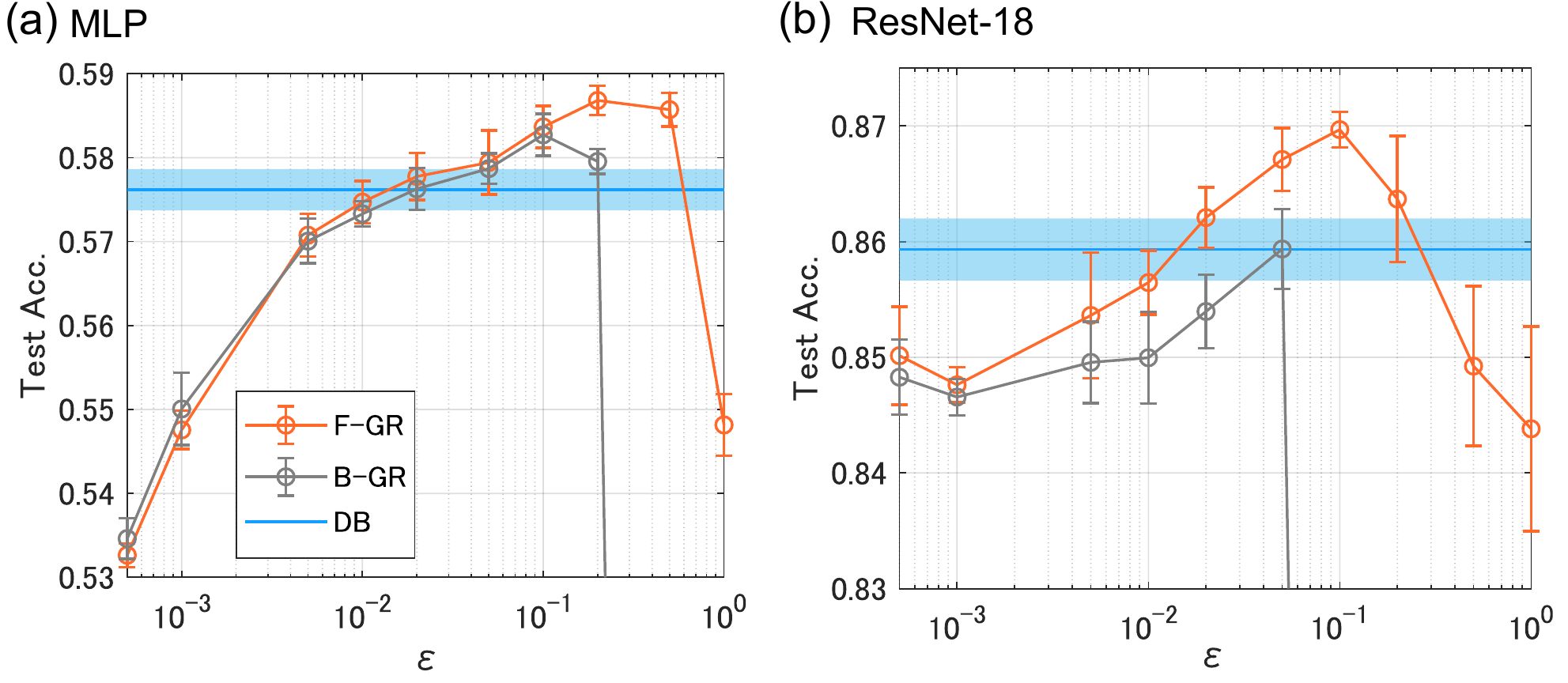}
\caption{Dependence of test accuracy on $\varepsilon$. We fixed $\gamma=0.5$ for MLP and $\gamma = 0.05$ for ResNet-18.}
\vspace{-5pt}
\label{fig3_fin}
\end{figure}

Figure \ref{fig4} shows the test accuracy of a 4-layer MLP and ResNet-18 trained by using SGD with GR on CIFAR-10. We trained the models in an exhaustive manner with various values for  $\varepsilon$ and $\gamma$ for each algorithm of the GR. 
For learning with F-GR, the model achieved the highest accuracy on relatively large ascent steps ($\varepsilon \sim 0.1$). Figure 
\ref{fig3_fin} shows a more quantitative visualization of the dependence on $\varepsilon$. F-GR with large $\varepsilon$ achieved better generalization performance than DB and B-GR. 
In Table \ref{tabS4}, we summarized the best test accuracy for all $\varepsilon$ and $\gamma$. This table also clarifies that the F-GR achieves the highest generalization performance.
We also confirmed that the same tendencies appeared in the grid search of ResNet-34 on CIFAR-100 (Figure \ref{figS3}). Furthermore, we confirmed in Figure \ref{figS_WRN} and Table \ref{tabS1} that F-GR performed better than B-GR and DB in the training of wide residual networks (WRN-28-10) on CIFAR-10 and CIFAR-100 with/without data augmentation. 

Note that in real training, the performance of F(B)-GR for a small $\varepsilon$ does not necessarily coincide with that of DB.
When the ascent step was too small, we observed numerical instability in the calculation of the gradient.  
It is also noteworthy that the best accuracy of F-GR was obtained close to the line of $\gamma = \varepsilon$. This line is closely related to SAM algorithm. We explain more details in Section \ref{Sec5_1}.
Overall, the experiments suggest that F-GR with a large ascent step is better to use for achieving higher generalization performance.

\subsection{Linear Model}
Although  previous work and our experiments in Section \ref{Sec3_3} indicate  improvements of prediction performance caused by GR, theoretical understanding of this phenomenon remains limited. Because the gradient norm itself eventually becomes zero after the model achieves a zero training loss, it seems challenging to distinguish the generalization capacity by simply observing the value of the gradient norm after training. In addition, our experiments clarified that the performance also depends on the choice of the algorithm and revealed that the situation is more complicated.

One approach to obtaining theoretical insight into empirical observation is to analyze them in a simple model.
First, let us consider a naive linear model $X \theta$, where $X$ denotes a data matrix and $\theta$ denotes training parameters. Interestingly,  the difference among GR algorithms {\it does not} appear in the linear model as follows. 
\begin{proposition}
Suppose a mean square error loss $\mathcal{L}(\theta)=\|X \theta-y\|^2/2$. Then, finite-difference GR has the same gradient as the original GR, that is,  
\begin{equation}
  \Delta R_F = \Delta R_B = \nabla R = X^\top X X^\top (X\theta-y), \label{eq7:0124}
\end{equation}
which is independent of $\varepsilon$. 
\end{proposition}
The derivation is straightforward. Note that
from the mean value theorem, the finite-difference GR is equivalent to
\begin{equation}
\Delta R_{F}(\varepsilon) =  \frac{1}{\varepsilon}\int_0^\varepsilon ds H(\theta_t+ s \nabla \mathcal{L}(\theta_t) )  \nabla \mathcal{L}(\theta_t). \label{eq6:0918}
\end{equation}
We can interpret the finite difference as taking an average of the curvature (Hessian) along the line of gradient update.  This includes $\Delta R_B(\varepsilon)$ for a negative $\varepsilon$ and $ \nabla R$ for $\varepsilon \rightarrow 0$. Because we have a constant Hessian $H=X^\top X$ for the above linear model, we immediately obtain (\ref{eq7:0124}) from (\ref{eq6:0918}).

Since the gradient is the same in the whole training, the eventual solution is also the same for for any $\varepsilon$. 
This result suggests that the difference of GR algorithms would be caused by some non-linearity of models.  
In the following, we show that the dependence on  GR algorithms  actually appears in a simple network model with non-linearity.

\subsection{Diagonal Linear Network (DLN) Model}
\label{Sec4_2f}

\subsubsection{Setting}
 A DLN is a solvable model proposed by \citet{woodworth2020kernel}. It is  a linear transformation of input $x \in \mathbb{R}^d$  defined as $\langle \beta, x \rangle$ where $\beta$ is parameterized in a non-linear way, that is,    
$\beta = w_+^2 -w_-^2$ with $w=(w_+,w_-) \in \mathbb{R}^{2d}$. Here, the square of the vector is an element-wise square operation. Suppose that we have $n$ training samples $(x^{(j)},y^{(j)})$ ($j=1,...,n$).   
The training loss is given by 
\begin{equation}
\mathcal{L}(w)=\frac{1}{4 n} \sum_{j=1}^n\left(\left\langle w_{+}^2-w_{-}^2, x^{(j)}\right\rangle-y^{(j)}\right)^2. \label{eq8:0927}
\end{equation}
Consider continual-time training dynamics $dw/dt = -\nabla \mathcal{L}$. We set an  initialization $w_{+}(t=0)=w_{-}(t=0)=\alpha_0$ \com{which is a $d$-dimensional vector} and whose entries are non-zero. We define a data matrix $X$ whose $i$-th row is given by $x^{(i)}$. 
\citet{woodworth2020kernel} found that interpolation solutions of usual gradient descent are given by 
\begin{equation}
\beta_{\infty}(\alpha)=\underset{\beta \in \mathbb{R}^d \text { s.t. } X \beta=y}{\arg \min } \phi_\alpha(\beta),
\label{eq10:0124}
\end{equation}
where $\alpha = \alpha_0$ and the potential function $\phi_\alpha$ is given by $\phi_\alpha(\beta)=\sum_{i=1}^d \alpha_i^2 q\left(\beta_i / \alpha_i^2\right)$ with $q(z)=2-\sqrt{4+z^2}+z \operatorname{arcsinh}(z / 2)$. 
For a larger scale of initialization $\alpha$,  this potential function becomes closer to L2 regularization as $\alpha_i^2 q(\beta_i/\alpha_i^2) \sim |\beta_i|^2$, which corresponds to the L2 min-norm solution of the lazy regime \citep{chizat2019lazy}. In contrast,
for a smaller scale of initialization $\alpha$, it becomes closer to L1 regularization as $\alpha_i^2 q(\beta_i/\alpha_i^2) \sim |\beta_i|$. In this way, we can observe a one-parameter interpolation between L1 and L2 implicit biases.  
Deep neural networks in practice acquire rich features depending on data structure and are believed to be beyond the lazy regime. Thus, obtaining an L1 solution by setting small $\alpha$ is referred to as the {\it rich regime} and desirable.  Previous work has revealed that effective values of $\alpha$ depend on algorithms. For example, $\alpha$ 
decreases by  
a larger learning rate in the discrete update \citep{nacson2022implicit}, SGD \citep{pesme2021implicit}, and SAM update \citep{andriushchenko2022towards}. It means that they have an implicit bias that chooses the L1 sparse solution in the rich regime.

\subsubsection{Results}

Now, we analyze a gradient flow with GR given by 
\begin{equation}
\frac{dw}{dt} = - \nabla \mathcal{L}(w) - \gamma  \Delta R_F(\varepsilon)  
\end{equation}
for a real value $\varepsilon \in \mathbb{R}$.  
Note that this expression includes not only the F-GR case but also the other cases as $\Delta R_B(\varepsilon) = \Delta R_F(-\varepsilon)$ and $\nabla R= \lim_{\varepsilon \rightarrow 0} \Delta R_F(\varepsilon)$. 
We find that the GR has implicit bias towards the rich regime, and moreover, the strength of the bias depends on the step size $\varepsilon$.

We use the following assumption:
\begin{assumption}
(i) the gradient dynamics converges to the interpolation solution satisfying $X\beta=y$, (ii) $\|w(t)\|$ has a constant upper \com{bound} independent of $\gamma$ and $\varepsilon$, (iii) for sufficiently small $\gamma$ and $\varepsilon$,  the integral of the training loss, i.e., \com{$\int_0^\infty \mathcal{L}(w(t)) dt$}, 
has a  constant upper bound $\overline{R}$ independent of $\gamma$ and $\varepsilon$. 
\end{assumption}
Assumption (i) is common among the studies of DLNs. Assumption (ii) is known to hold under a certain condition identified by \citet{nacson2022implicit}.
 Assumption (iii) is related to the convergence speed of training dynamics and a sufficient condition that the dynamics converge to the interpolation solution. See Section \ref{SecA2} for more details. We find the following:
\begin{theorem}
Under Assumption 4.2, for sufficiently small $\gamma$, interpolation solutions are given by $\beta_{\infty}(\alpha_{GR})$ with
\begin{equation}
\alpha_{GR} = \alpha_0 \circ \exp(- \gamma( c_0+ \varepsilon c_1 + \varepsilon^2 c_2  )+ \com{\mathcal{O}(\gamma^2)} ), \label{eq10:0927}
\end{equation}
where
\begin{align}
c_0 &=  \int_0^\infty (X^\top (X\beta(s)-y))^2 ds/n^2, \\
    c_1 &=(X^\top (X\beta(t=0)-y))^2/2n^2, \label{eq5:0901}
\end{align}
and $c_2$ is a $d$-dimensional vector.
\end{theorem}
The proof is given in Section \ref{SecA1}. 
Note that $c_0$, $c_1$ and $c_2$ are $d$-dimensional vectors and $\circ$ denotes an entry-wise product.  
This theorem clarifies the dependence of the solution on the step size $\varepsilon$. The positive $c_0$ term is a factor that makes the solution biased towards the rich regime for all $\varepsilon$. The problem is how $\varepsilon c_1$ and $\varepsilon^2 c_2$ terms determine  the eventual value of $\alpha_{GR}$.   
First, let us neglect the $\varepsilon^2 c_2$ term by taking a sufficiently small $|\varepsilon|$. Then, we can see that $\alpha_{GR}$ gets smaller than $\alpha_0$ for $\varepsilon>0$ because of the positivity of $c_0$ and $c_1$. In other words, F-GR provides an implicit bias towards the rich regime. 
In contrast, for $\varepsilon<0$, the $\varepsilon c_1$ term takes a negative value and this suggests that B-GR is not necessarily biased towards the rich regime. 

Next, for a more quantitative evaluation, we provide an upper bound of $\alpha_{GR}$ for F-GR: 
\begin{proposition}
Suppose the $i$-th entry of $c_1$ is non-zero, i.e., $c_{1,i}>0$.  
Under Assumption 4.2, by taking small positive $\varepsilon$ and $\gamma$ satisfying $0<\varepsilon\leq \varepsilon'$ and $0<\gamma\leq\gamma'$ for some constants $\varepsilon'$ and $\gamma'$, we have  
\begin{equation}
\alpha_{GR,i} \leq \alpha_{0,i} \exp (-\gamma \varepsilon c_{1,i}/2).   
\end{equation}
\end{proposition} 
It is highly likely for $c_1$ to take non-zero values because $c_1$ is determined by initialization and we usually have $X\beta(t=0) \neq y$.
The deviation is shown in  Section \ref{Sec_A3} and detailed definitions of constants $\gamma'$ and $\varepsilon'$ are given in Eqs. (\ref{gamma_dash},\ref{var_dash}).
The proposition clarifies that F-GR has an implicit bias to select the L1 solution, that is, the rich regime because $\alpha$ is always smaller than $\alpha_0$. 
In the same way, for $\varepsilon<0$ and a sufficiently small $|\varepsilon|$,  we can immediately find
\begin{equation}
\alpha_{GR,i} \geq \alpha_{0,i} D^\gamma \exp (\gamma |\varepsilon| c_{1,i}) \label{eq16:0124}
\end{equation}
where $D$ is a constant scalar. This inequality reveals that B-GR has an increasing lower bound for a larger $|\varepsilon|$. It suggests that B-GR has an implicit bias towards the lazy regime.

\begin{figure}
\centering
    \includegraphics[width=0.65\textwidth]{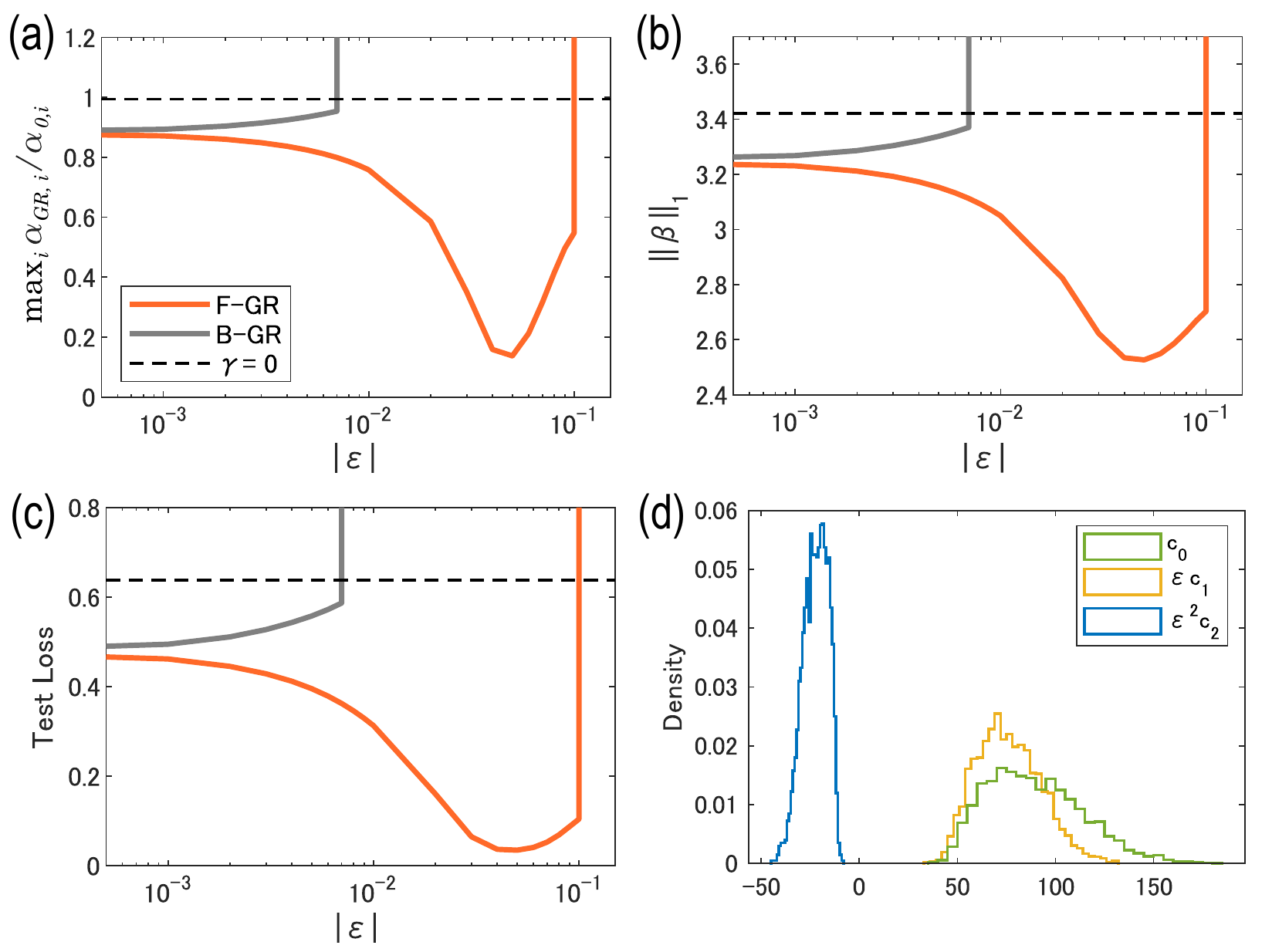}
\caption{Experimental results of DLNs trained by gradient descent with F-GR/B-GR ($\gamma=0.02$). (a) Test loss, (b) The largest $\alpha_{GR,i}$ over $i=1,...,d$, (c)
L1 norm of the solutions, (d) Distribution of exponents after training with F-GR ($\varepsilon =0.05$). }
\vspace{-5pt}
\label{fig5_DLN}
\end{figure}

 Figure \ref{fig5_DLN} confirms our theory by numerical experiments.
As in previous work, we trained DLNs on the synthetic data of a sparse regression problem, where $x^{(j)} \sim \mathcal{N}(\mu 1,\sigma^2 I)$ and $y^{(j)} \sim \mathcal{N}\left(\left\langle \beta^*, x^{(j)}\right \rangle, 0.01\right)$, and where  $\beta^*$ is $k^*$-sparse with non-zero entries equal to $1/\sqrt{k^*}$ ($d=100$ and $n=50$).  
Following \cite{nacson2022implicit}, we chose $\mu=\sigma^2=5$, where the parameter norm $a(t)$ is suppressed and assumption (ii) is expected to hold. 
We initialized parameters by $\alpha_{0,i} \sim \mathcal{N}(0, 0.01)$.   
The solid lines show the results of actual gradient descent training with F-GR or B-GR. The dashed lines show the results without GR. Other technical details including  methods
to empirically estimate $\alpha_{GR}$ and $c_i$ are summarized in  Section \ref{Sec_exp_DLN}. 

As the ascent step increased,  the models trained by F-GR
initially achieved 
 smaller $\alpha_{GR}$ (Figure \com{\ref{fig5_DLN}(a)}) and sparser solution (Figure \com{\ref{fig5_DLN}(b)}) 
as is expected from our theory. This led to 
better generalization (Figure \com{\ref{fig5_DLN}(c)}).  
Note that the improvement of generalization caused by the sparse solution is widely observed in the studies of other learning algorithms \cite{nacson2022implicit}.
 After the step size increased to some degree, $\alpha_{GR}$ 
 increased slightly, and then the training dynamics exploded for too large $\varepsilon$. This increase of $\alpha_{GR}$ is also consistent with our theory because we empirically observed negative $\varepsilon^2 c_2$ terms (Figure \ref{fig5_DLN}(d)) and they can make the $\alpha_{GR}$ increased as in Eq. (\ref{eq10:0927}). 
 It is noteworthy that the performance of 
 more realistic neural networks (Figure \ref{fig3_fin}) showed qualitatively similar behavior, where the best generalization performance was achieved by the F-GR with a large ascent step.
 In Figure S.4, we also present the largest eigenvalue of the Hessian (\ref{S47:0928}), computed after training. As the ascent step size increased, F-GR chose flatter minima. This is also consistent with empirical observations of GR \cite{barrett2020implicit}. 
For B-GR, we can see that $\alpha_{GR}$ increased as $|\varepsilon|$ increased, as is expected from the implicit bias to the lazy regime (\ref{eq16:0124}).

\section{Implicit Finite-Difference GR}

So far, we have obtained a better understanding of explicit GR, especially, finite-difference GR.
Here, we show that the  GR has hidden connections to other gradient-based learning methods. 
We recall that the finite-difference GR is composed of both gradient ascent and descent steps. This computation makes it essentially related to two other learning methods similarly composed of both gradient ascent and descent steps: the flooding method and the SAM algorithm.

\subsection{Flooding}
\label{Sec5_2}
The flooding method \citep{ishida2020we} is a learning algorithm composed of both gradient ascent and descent steps. 
 Its update rule is given by 
\begin{align}
\theta_{t+1} = \theta_t - \eta \mathrm{Sign}(\mathcal{L}-b) \nabla \mathcal{L} 
\end{align}
for a constant $b>0$, referred to as the flood level. When the training loss becomes lower than the flood level, the sign of the gradient is flipped and the parameter is updated by gradient ascent. Therefore, the flooding causes the training dynamics to continue to wander around $\mathcal{L}(\theta) \sim b$, and its gradient continues to take a non-zero value.
This would seem a kind of early stopping, but previous work empirically demonstrates that flooding performs better than naive early stopping and finds flat minima.
For simplicity, let us focus on the gradient descent for a full batch. The following theorem clarifies a hidden mechanism of  flooding.
\begin{theorem}
Consider the time step $t$ satisfying  $\mathcal{L}(\theta_{t})<b$ and  $\mathcal{L}(\theta_{t+1})>b$. Then, the flooding update from $\theta_t$ to $\theta_{t+2}$ is equivalent to the gradient of the F-GR with  $\varepsilon=\gamma=\eta$:
\begin{align}
\theta_{t+2} &=  \theta_t - \eta^2 \frac{ \nabla \mathcal{L}( \theta_t + \eta \nabla \mathcal{L}(\theta_t)) - \nabla \mathcal{L}(\theta_t)}{\eta}. 
\end{align}
Similarly, for $\mathcal{L}(\theta_{t})>b$ and  $\mathcal{L}(\theta_{t+1})<b$, the flooding update is equivalent to the gradient of the B-GR.
\end{theorem}
Although its derivation is quite straightforward (see Section \ref{Sec_B}), this essential connection between finite-difference GR and flooding has been missed in the literature.
\citet{ishida2020we}  conjectured that flooding causes a random walk on the loss surface and this would contribute to the search for flat minima in some ways. Our result implies that the dynamics of flooding are not necessarily random and it can actively search the loss surface in a direction that decreases the GR. This is consistent with the observations that the usual gradient descent with GR finds flat minima \citep{barrett2020implicit,zhao22i}.

\begin{figure}
\centering
    \includegraphics[width=0.8\textwidth]{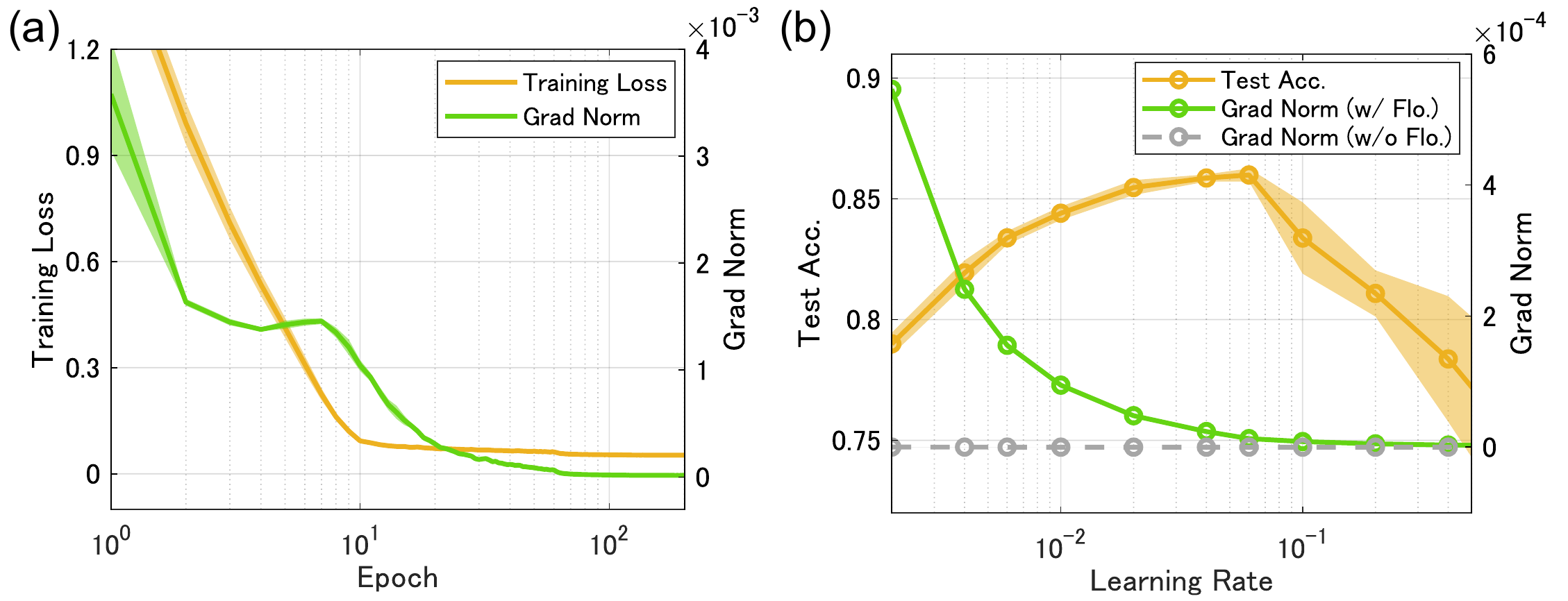}
\caption{Flooding decreases the gradient norm, as expected by theory. (a) Training dynamics of flooding with $b=0.05$. (b) Test accuracy and gradient norm after the training. }
\vspace{-5pt}
\label{fig6}
\end{figure}

Figure \ref{fig6} empirically confirms that the flooding method decreases the gradient norm $R(\theta)$.  We trained ResNet-18 on CIFAR-10 by using flooding. 
Figure \ref{fig6}(a) shows that at the beginning of the training, the training loss decreased in the usual way because the loss was far above flood level $b$. Around the 10th epoch, the loss value became sufficiently close to the flood level for the decrease in the loss to slow \com{(Figure \ref{figS6})}. Then, the flooding update became dominant in the dynamics the gradient norm began to decrease. Figure \ref{fig6}(b) demonstrates that the gradient norm of the trained model decreased as the initial learning rate increased. This is consistent with Theorem 5.1 because the theorem claims that the larger learning rate induces the larger regularization coefficient of the GR $\gamma=\eta$. In contrast, naive SGD training without flooding always reaches an almost zero gradient norm regardless of the learning rate. Thus,  the change in the gradient norm depending on the learning rate is specific to flooding and implies that it implicitly performs GR through the finite difference computation.

\subsection{SAM}
\label{Sec5_1}
Finally, let us give a remark on a connection with SAM.
The SAM algorithm was derived from the minimization of a surrogate loss $\max_{\|\varepsilon \| \leq \rho} \mathcal{L}(\theta+ \varepsilon)$ for a fixed $\rho>0$, and has achieved the highest performance in various models \citep{foret2020sharpness}. After some heuristic approximations,  
its update rule  reduces to iterative gradient ascent and descent steps: $\theta_{t+1}=\theta_t - \eta \nabla \mathcal{L}(\theta')$ with $\theta' = \theta_t + \varepsilon_t \nabla \mathcal{L} (\theta_t) $ and $ \varepsilon_t =\rho/\|\nabla \mathcal{L}(\theta_t)\|$. 
Under a specific condition, the SAM update can be seen as gradient descent with F-GR. Let us consider  time-dependent regularization coefficient $\gamma_t$ and ascent step  $\varepsilon_t$. Then, for   $\gamma_t = \varepsilon_t$, the gradient descent with F-GR becomes equivalent to the SAM update:
\begin{equation}
  \nabla \mathcal{L}(\theta) + \frac{\gamma_t}{\varepsilon_t} ( \nabla\mathcal{L}(\theta') -  \nabla \mathcal{L}(\theta) ) =  \nabla \mathcal{L}(\theta').
\end{equation}
A similar equivalence has been pointed out in \citet{zhao22i} which supposes a non-squared gradient norm and $ \varepsilon_t =\rho/\|\nabla \mathcal{L}(\theta_t)\|$ naturally appears. 
Suppose the SAM update without the gradient normalization for simplicity, that is, $\varepsilon_t = \rho$. 
This simplified SAM update was analyzed on DLNs in \citet{andriushchenko2022towards}. We can recover their expression of $\alpha$ by setting a sufficiently small $\gamma = \varepsilon$ and neglecting $c_1$ and $c_2$ terms in Theorem 4.3. 

It will be curious to identify any optimal setting of ($\varepsilon,\gamma$) for generalization performance although we remain it as future work. 
In Figure \ref{fig4}, we empirically observed the optimal setting for generalization was very close to or just on the line $\gamma = \varepsilon$. In contrast, our experiments on DLN (Figures \ref{fig5_DLN} \& \ref{figS5_DLN}) and the previous study \citet{zhao22i} demonstrated that the optimal setting was not  on $\gamma = \varepsilon$, and thus combining the ascent and descent steps would be still promising.

\section{Discussion}
 
 This work presented novel practical and theoretical insights into GR. The finite-difference computation is effective in the sense of both reducing computational cost and improving performance. Theoretical analysis supports the empirical observation that 
 the forward difference computation has an implicit bias that chooses potentially better minima depending on the size of the ascent step. 
 Because deep learning requires large-scale models, it would be reasonable to use learning methods only composed of  first-order descent or ascent gradients. The current work suggests that the F-GR is a promising direction for further investigation and could be extended for our understanding and practical usage of gradient-based regularization.

We suggest several potentially interesting research directions.
 From a broader perspective, we may regard finite-difference GR, SAM, and flooding as a single learning framework composed of iterative gradient ascent and descent steps. It would be interesting to investigate if there is optimal combination of these steps for further improving performance. 
 Related to the combination between the gradient descent and ascent, although we fixed the ascent step size as a constant, a step size decay or any scheduling could enhance the performance further. For instance, \citet{zhuang2022surrogate} used a time-step dependent ascent step to achieve high prediction performance for SAM.
These advanced topics could be interesting for developing further efficient algorithms or regularization methods. 

It will also be interesting to explore any theoretical clarification beyond the scope of DLNs. Although a series of analyses in DLNs enable us to explore the implicit bias for selecting global minima, it assumes global convergence and avoids an explicit evaluation of convergence dynamics.
Thus, it would be informative to explore the convergence rate or escape from local minima in other solvable models or a more general formulation if possible.
Constructing generalization bounds would also be an interesting direction. Some theoretical work has proved that regularizing first-order derivatives of the network output controls the generalization capacity \citep{ma2021linear}, and such derivatives are included in the gradient norm as a part.
We expect that the current work will serve as a foundation for further developing and understanding regularization methods in deep learning.

\bibliography{test}
\bibliographystyle{iclr2023_conference}

\newpage

{\bf \huge Appendices}

\setcounter{section}{0}
\renewcommand{\thesection}{\Alph{section}}
\renewcommand{\theequation}{S.\arabic{equation} }
\renewcommand{\thefigure}{S.\arabic{figure} }
\renewcommand{\thetable}{S.\arabic{table} }

\setcounter{equation}{0}
\setcounter{figure}{0}

\section{Computational Aspect of GR}
\label{secDB}

\subsection{Pseudo-code and implementation}
\label{pcode}
In the experiments on benchmark datasets, we computed the GR term in each mini-batch of SGD update. The pseudo-code for F-GR is given in Algorithm 1. 
The double backward computation is implemented as shown in Listing 1. 

\begin{algorithm}[H]
\begin{algorithmic}[1]
\caption{Learning with F-GR}\label{alg:cap}
\renewcommand{\algorithmicensure}{\textbf{Input:}}
\renewcommand{\algorithmicrequire}{\textbf{Define:}}
\ENSURE  $\text{mini-batches} \{B_1, ...,B_K\}$ 
\WHILE{SGD update}
\IF{$i$-th mini-batch}
\STATE $\Delta \mathcal{L} \gets \nabla \mathcal{L}(\theta;B_i)$
\STATE $\theta' \gets \theta + \varepsilon \Delta \mathcal{L}$ 
\STATE $\Delta \mathcal{L}' \gets \nabla \mathcal{L}(\theta';B_i)$
\STATE $\Delta R \gets  (\Delta \mathcal{L}'-\Delta \mathcal{L})/\varepsilon $ 
\STATE $\theta \gets \theta - \eta (  \Delta \mathcal{L} + \gamma  \Delta R)$ 
\ENDIF
\ENDWHILE
\end{algorithmic}
\end{algorithm}

\begin{lstlisting}[language=Python, caption=Implementation of DB in PyTorch.]
...
loss.backward(create_graph=True) #backpropagation of original loss
loss_DB = (gamma/2)*sum([torch.sum(p.grad**2) for p in model.parameters()]) #computing GR term
loss_DB.backward() #backpropagation of GR term    
optimizer.step()
...

\end{lstlisting}

\subsection{Evaluation on the number of Matrix Multiplication}
\label{S_matmul}

We represent an $L$-layer fully connected neural network with a linear output layer 
by $A_{l}=\phi (U_{l})$, $U_{l}= W_{l} A_{l-1}$
for $l=1,...,L$. 
We define the element-wise activation function by $\phi(\cdot)$ and weight matrix by $W_{l}$. For simplicity, we neglect bias terms. Note that we have multiple samples $A_{0}$ (within each minibatch)
as an input and $W_{l} A_{l}$ requires a matrix-matrix product. 
Therefore, the forward pass requires $L$ matrix multiplication. 
Next, let us overview usual backpropagation on
the forward pass $\{A_0 \rightarrow A_1 \rightarrow \cdots \rightarrow A_L  \}$. We can express the backward pass as $B_{l} = \phi'(U_{l}) \circ  (W_{l+1}^\top B_{l+1})$, where the backward signal $B_l$ corresponds to $\partial \mathcal{L}/\partial U_l$ ($l=1,...,L-1$).
Then, the backward pass requires $L-1$ matrix-matrix multiplication between weights $W$ and backward signals $B$.
In addition, we need to compute the gradient $\partial \mathcal{L}/\partial W_l = B_l A_{l-1}^\top$ for $\nabla \mathcal{L}$ and this is also a matrix-matrix multiplication. Alter all, we need $3L-1$ matrix multiplication for $\nabla \mathcal{L}$. 

\noindent 
{\bf Finite difference computation: }  
$\nabla \mathcal{L}(\theta')$ requires the same number of matrix multiplication as the normal backpropagation. Therefore, $\nabla \tilde{\mathcal{L}}$ requires $6L-2$. For a sufficiently deep network, this is $\sim 6L$.

\noindent 
{\bf Double Backward computation: }
 Let us denote $\partial \mathcal{L}/\partial W_l$ by $G_l$. 
 Figure \ref{fig1} represents the forward pass for computing the gradient of GR. 
Note that the upper part of this graph, i.e., $\{A_0 \rightarrow A_1 \rightarrow \cdots \rightarrow B_L  \rightarrow \cdots \rightarrow B_1 \}$, is well-known in double backpropagation of $\nabla B_{1}$ for the input-Jacobian regularization. As explained in \citet{drucker1992improving}, the computation of $\nabla B_{1}$ is equivalent to apply backpropagation to this upper part of the graph. GR requires additional $L$ nodes for $G_l$. 
Note that when we have a forward pass with matrix multiplication, its backward computation requires two matrix multiplications. That is, when a node of the forward pass $S$ is a function of the matrix $X$ given by $X=UV$, we need to compute $\partial S/ \partial U = (\partial S/ \partial X) V $ and $\partial S/ \partial V = U(\partial S/ \partial X) $ in the backpropagation. In addition, we do not need to compute the derivative of $A_0$. After all, we need  $2\times(3L-1)-2 = 6L-4$  for the $\nabla R$.  Since we also compute the gradient of the original loss $\nabla \mathcal{L}$, we need $9L-5$. For a sufficiently deep network, this is $\sim 9L$.

\section{Details of Experiments}
\label{Sec_C2}

\subsection{Computational Aspect}

\noindent
{\bf Figure 2:} We trained MLP (width 512) and ResNet  on CIFAR-10  by using SGD with GR. We used  Rectified Linear Units (ReLUs) for activation functions, and set batch size 256, momentum 0.9, initial learning rate 0.01 and used a step decay of the learning rate (scaled by 5 at epochs 60, 120, 160), $\gamma=\varepsilon = 0.05$ for GR. We showed the average and standard deviation over 5 trials of different random initialization. 

\noindent
{\bf Figure S.1:}
This figure shows the trajectories of the original training loss $\mathcal{L}$ during the training. Its setting is the same as in Figure \ref{fig2}. 
We observed that learning with F-GR could make the loss decrease faster than DB in the sense of convergence rate (i.e., the number of epochs). This means that the loss converges even faster in wall time.  

\begin{figure}[h!]
\centering
    \includegraphics[width=0.7\textwidth]{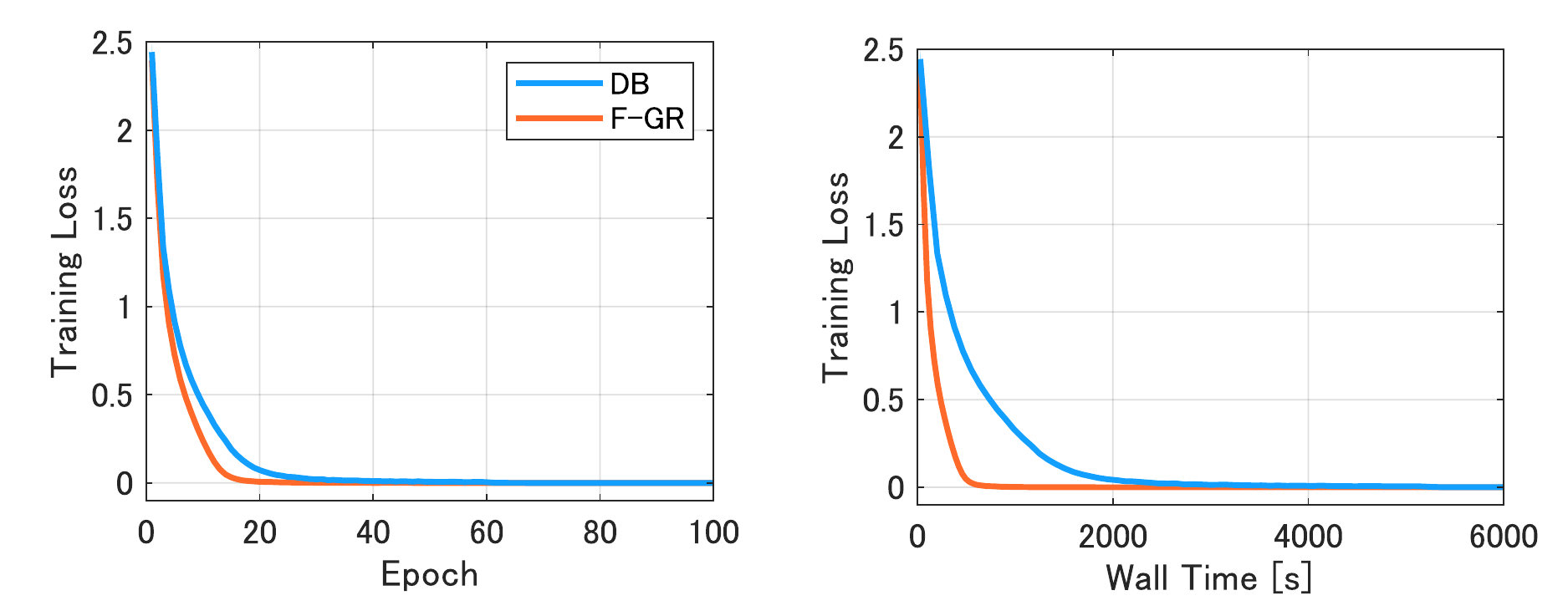}
\caption{Training dynamics in ResNet-18 on CIFAR-10. Learning with F-GR is much faster in wall time.
}
\vspace{-5pt}
\label{figS1}
\end{figure}

\subsection{Generalization Performance}

\subsubsection*{MLP and ResNet}

\noindent
{\bf Figure 3:}  We trained (a) 4-layer MLP  and (b) ResNet-18 on CIFAR-10 by using SGD with GR. We trained the models with various hyper-parameters $\varepsilon=\{10^{-5},5\times 10^{-5}, ..., 0.5, 1\}$ and $\gamma =\{10^{-4}, 2 \times 10^{-4}, 5 \times 10^{-4}, 10^{-3}, ..., 1, 2,5\}$. The other settings are the same as in Figure \ref{fig2}. We set batch size 128, weight decay 0.0001, and used no other regularization technique or data augmentation.  Table \ref{tabS4}
shows the highest average test accuracy among all settings of $\gamma$ and $ \varepsilon$. 

\begin{table}[h]
  \caption{Summary of the highest test accuracy in the grid search of ($\varepsilon,\gamma$)  shown in Figure \ref{fig4}.  }
  \centering
  \begin{tabular}{lll}
    \toprule
   & MLP  & ResNet-18   \\ 
    \midrule
   F-GR & {\bf 58.6}{\tiny $\pm$ 0.2} & {\bf 87.0} {\tiny $\pm$ 0.2}\\ 
   B-GR & {58.3}{\tiny $\pm$ 0.2} & 86.2 {\tiny $\pm$ 0.3} \\
   DB    & {57.6}{\tiny $\pm$ 0.2} & 86.3 {\tiny $\pm$ 0.3}  \\   
   \bottomrule
  \end{tabular}
  \label{tabS4}
\end{table}

\noindent
\com{{\bf Figure 4:} To see the difference among algorithms in more detail, we show test accuracy along $\varepsilon$ axis with a fixed $\gamma$ of the grid search shown in Figure 3.  Each line represents the average and standard deviation over 5 trials of different random initialization. 
We fixed $\gamma=0.5$ for MLP and $\gamma = 0.05$ for ResNet-18.  This means that the objective function is the same among different algorithms. Nevertheless, the eventual performance is different. For a large $\varepsilon$, F-GR
achieves the higher test accuracy than DB beyond one standard deviation. For such a large $\varepsilon$, F-GR also performs better than B-GR. }

\noindent
{\bf Figure S.2:} We trained ResNet-34 on CIFAR-10. (a) This figure shows the same grid search as is shown in Section \ref{Sec3_3}. 
 The result is consistent with those in MLP and ResNet-18 (Figure \ref{fig4}). Learning with F-GR achieved the highest accuracy for large ascent steps. In addition, it was better than the highest accuracy of DB. The best test accuracy was $(\text{F-GR},\text{B-GR},\text{DB})=(59.9,58.6,59.5)\pm(0.5,0.4,0.5)$. 
(b) This figure shows test accuracy along the $\varepsilon$ axis with a fixed $\gamma=0.05$.  Each line represents the average and standard deviation over 5 trials of different random initialization. As is similar to Figure 4, F-GR achieves the highest test accuracy.

\begin{figure}[t]
\centering
    \includegraphics[width=0.8\textwidth]{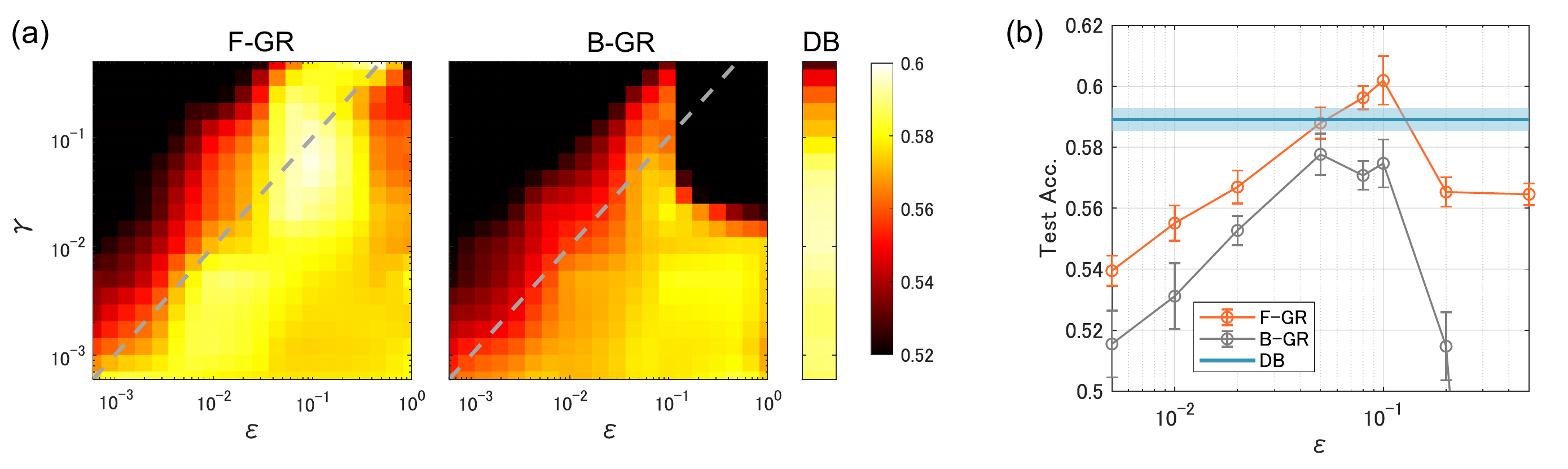}
\caption{Learning with different GR algorithms in ResNet-34 on CIFAR-100. 
(a) The color map shows the average test accuracy over 5 trials. Gray dashed lines indicate $\gamma=\varepsilon$.  (b) Case of $\gamma=0.05$.
}
\vspace{-5pt}
\label{figS3}
\end{figure}

\subsubsection*{WideResNet}

 \noindent 
{\bf Table S.2 and Figure S.3}: We trained WideResNet-28-10 (WRN-28-10) with $\gamma = \{0,  10^{-4}, 10^{-3}, 10^{-2}, 10^{-1}\}$.
For F-GR and B-GR, we set $\epsilon = \{0.01, 0.02, 0.05, 0.1, 0.2, 0.5\}$. 
We computed the average and standard deviation over 5 trials of different random initialization. 
We used crop and horizontal flip as data augmentation, cosine scheduling with an initial learning rate of 0.1, and set momentum 0.9, batch size 128, and weight decay 0.0001. Table S.2 reported the best average accuracy achieved over all the above combinations of hyper-parameters. R-GR achieves the highest test accuracy in all cases. 
Figure S.3 shows the test accuracy with $\gamma=0.1$ for F/B-GR and the highest test accuracy of DB over all $\gamma$. It clarifies that the F-GR achieves the highest accuracy for large $\varepsilon$ and performs better than B-GR and DB.

\begin{figure}
\centering
    \includegraphics[width=1\textwidth]{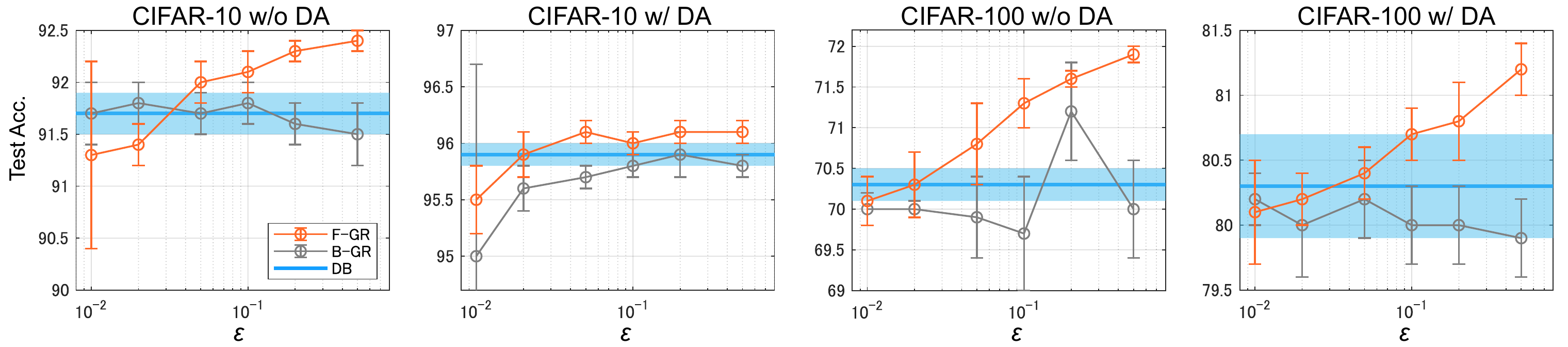}
\caption{ Learning with different GR algorithms in WideResNet-28-10 ($\gamma=0.1$). 
}
\vspace{-5pt}
\label{figS_WRN}
\end{figure}

\begin{table}[h]
  \caption{Test accuracy of WRN-28-10 shows that F-GR performs better. We trained the models with/without data augmentation (DA). }
  \centering
  \begin{tabular}{lllll}
    \toprule
    &  \multicolumn{4}{c}{WRN-28-10}\\
    &  \multicolumn{2}{c}{CIFAR-10} & \multicolumn{2}{c}{CIFAR-100} \\
    & w/o DA  & w/ DA &  w/o DA & w/ DA   \\ 
    \midrule
   F-GR & {\bf 92.4}{\tiny $\pm$ 0.1} & {\bf 96.1}{\tiny $\pm$ 0.1} & {\bf 71.9}{\tiny $\pm$ 0.1} & {\bf 81.2}{\tiny $\pm$ 0.2} \\ 
   B-GR  &  {91.9}{\tiny $\pm$ 0.1} & {95.9}{\tiny $\pm$ 0.1} & {71.2}{\tiny $\pm$ 0.6}& {80.2}{\tiny $\pm$ 0.2}  \\ 
   DB    & {91.7}{\tiny $\pm$ 0.2} & {95.9}{\tiny $\pm$ 0.1} & {70.3}{\tiny $\pm$ 0.2} & {80.3}{\tiny $\pm$ 0.4} \\  
   \bottomrule
  \end{tabular}
  \label{tabS1}
\end{table}

\subsection{Diagonal Linear Network}
\label{Sec_exp_DLN}

\noindent
{\bf Figures 4 and S.4:} We generated synthetic data by $x^{(j)} \sim \mathcal{N}(\mu 1,\sigma^2 I)$ and $y^{(j)} \sim \mathcal{N}\left(\left\langle \beta^*, x^{(j)}\right \rangle, 0.01\right)$.  $\beta^*$ is $k^*$-sparse with non-zero entries equal to $1/\sqrt{k^*}$.  
We set $d=100$, $n=50$, $\mu=\sigma^2=5$, $\gamma = 0.02$ and initialization $\alpha_{0,i} \sim \mathcal{N} (0,0.01)$. We trained the models by the discrete update of gradient descent with a small learning rate $\eta=0.001$. We trained the models until the training loss $\mathcal{L}$ became lower than $10^{-8}$.
We showed the average of 40 trials with different seeds. 

\begin{figure}[h]
\centering
    \includegraphics[width=0.7\textwidth]{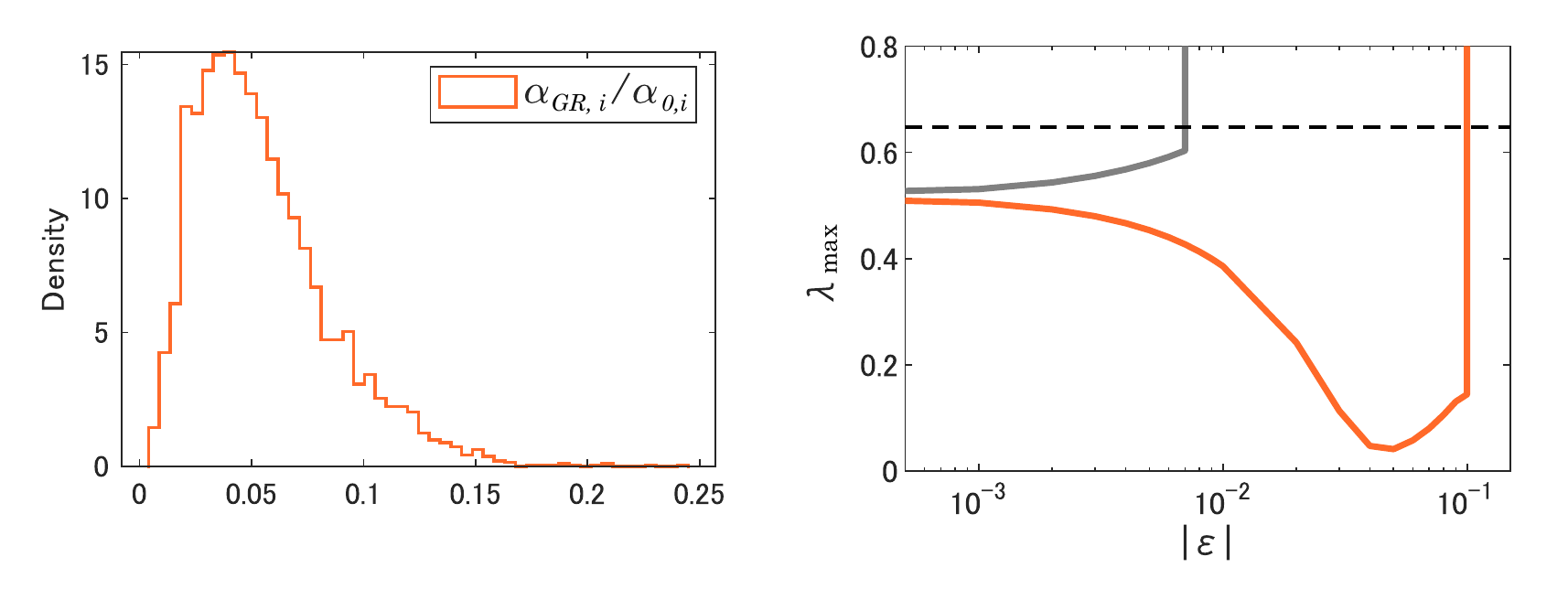}
\caption{Supplementary figures for the experiments of DLN. (Left) The largest eigenvalue of Hessian. (right) Density of $\alpha_{GR,i}/\alpha_{0,i}$ ($\gamma=0.02, \varepsilon=0.05$).} 
\vspace{-5pt}
\label{figS4}
\end{figure}

\begin{figure}[h]
\centering
    \includegraphics[width=1\textwidth]{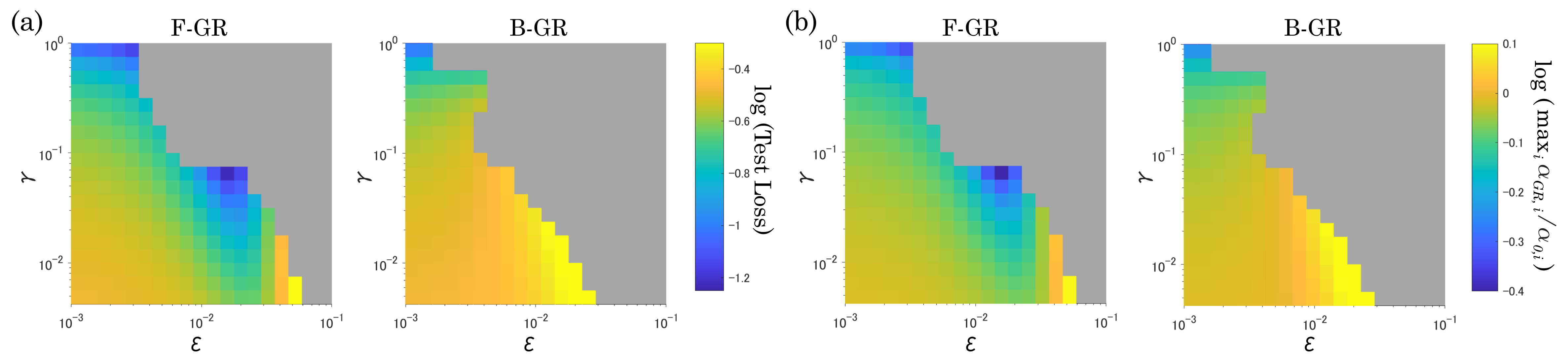}
\caption{Training of DLNs by GR with various $\gamma$ and $\varepsilon$. (a) Test Loss. (b) The largest $\alpha_{GR,i}$ over $i=1,...,d$. 
 Training dynamics exploded in the gray area. } 
\vspace{-5pt}
\label{figS5_DLN}
\end{figure}

\begin{figure}[h]
\centering
    \includegraphics[width=0.4\textwidth]{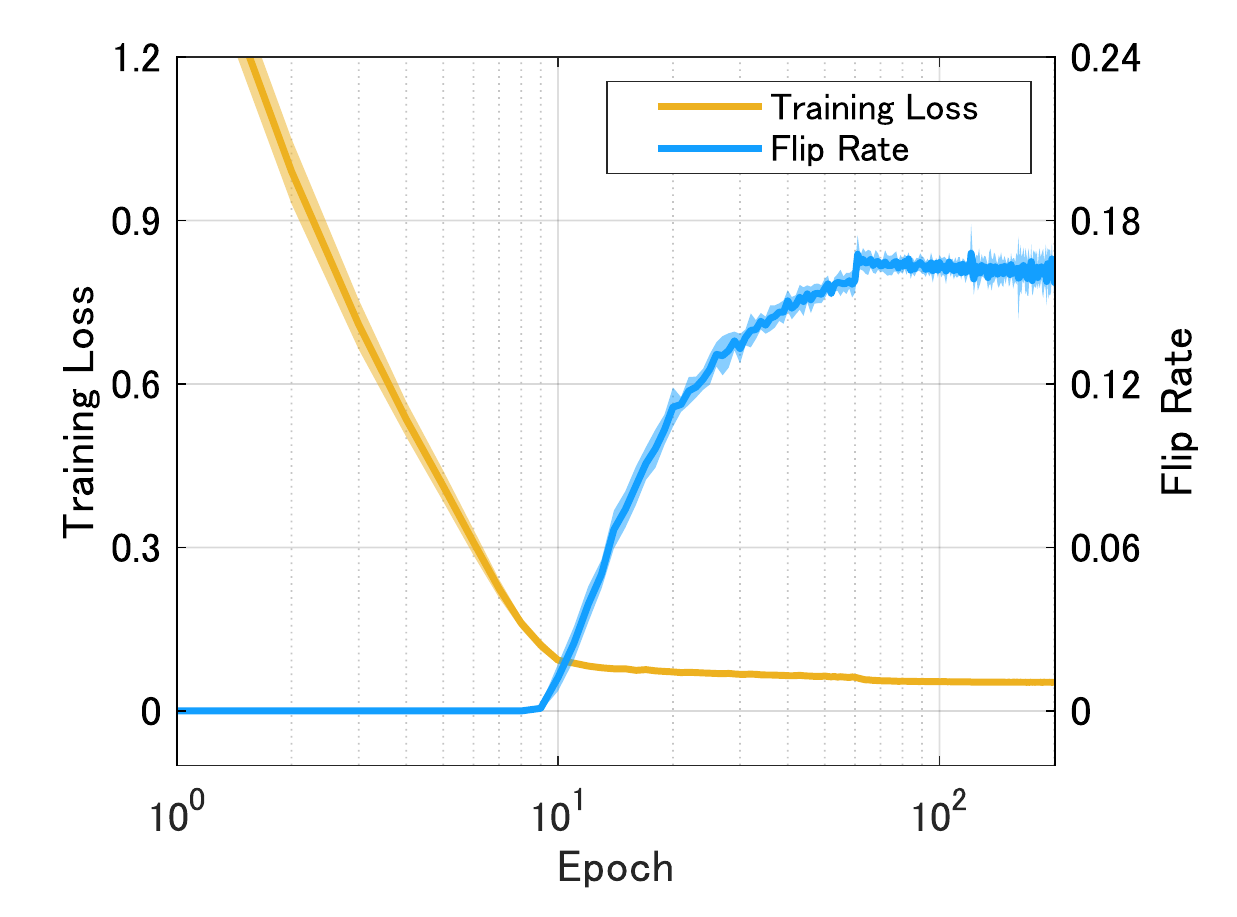}
\caption{\com{Flip rate of flooding with $b=0.05$.}} 
\vspace{-5pt}
\label{figS6}
\end{figure}

In numerical experiments of training DLNs, we can estimate $\alpha_{GR}$ without explicitly evaluating $\Psi$. From Eq. (\ref{S6:0923}), we have
\begin{align}
    w_+(\infty) \circ w_-(\infty) &= \alpha_0^2 
    \circ \exp \left(-\frac{\gamma}{n^2} \Psi \right). \label{S1:0124}
\end{align}
This leads to the following formula: 
\begin{equation}
    \alpha_{GR} = \sqrt{w_+(\infty) \circ w_-(\infty)}. 
\end{equation}
Thus, we can estimate $\alpha_{GR}$ by using the parameters eventually obtained by gradient dynamics. We computed $\alpha_{GR}$ shown in Figure \ref{fig5_DLN}(a) by using this formula. We can also obtain the density distribution of $\alpha_{GR,i}$ as is shown in Figure \ref{figS4}(left).

In Figure \ref{fig5_DLN}(b), we plotted the density distributions of exponents $c_0$, $c_1$ and $c_2$ after the training. The exponent $c_1$ is determined at initialization and is easy to compute.   
As is shown in Section \ref{SecA1}, we have $c_0 = \Psi_0/n^2$ and $c_2=\Psi_2/n^4$ where $\Psi_0$ and $\Psi_2$ are obtained by integrals over time (\ref{S18}). 
We numerically estimated them by taking the summation over the steps of gradient descent. For instance, we computed $c_0 \approx \sum_{t=0} (X^\top r(t))^2 \eta$ where $\eta$ is the learning rate of gradient descent. 

Figure \ref{figS4}(right) shows the largest eigenvalue of the Hessian. 
For the MSE loss of the DLN, the Hessian is given by 
\begin{equation}
H=\frac{1}{n}\left( \text{diag}(\tilde{X}^\top r) + 2\text{diag}(w)\tilde{X}^\top \tilde{X} \text{diag}(w) \right).
\end{equation}
At the interpolation solution, we have
\begin{equation}
 H=   \frac{2}{n}\text{diag}(w)\tilde{X}^\top \tilde{X} \text{diag}(w) \label{S47:0928}.
\end{equation}

 \noindent
{\bf Figure S.5}: We trained DLNs with various $\varepsilon$ and $\gamma$ in the same setting as in Figure 5. The color map shows the average over 10 trials. One can see that the test loss is correlated very well with $\alpha_{GR}$. While the test loss and $\alpha_{GR}$ could decrease as $\varepsilon$ increases in F-GR, they increased in B-GR. This behavior is consistent with our theory.

\subsection{Flooding Method}
    
 \noindent
\com{{\bf Figure S.6}: This figure confirms at which epoch the training loss started to get close to the flood level. The experimental setting is the same as in Figure \ref{fig6}. The blue line shows a flip rate, that is, the ratio of how many times the training loss gets smaller than the flood level during each epoch. Around the 10th epoch, the training loss started to reach the flooding level and the gradient norm also started to decrease. } 

\section{Analysis in Diagonal Linear Networks}

\subsection{Proof of Theorem 4.3}
\label{SecA1}
\subsubsection{Interpolation Solutions between L1 and L2 regularization}

We consider the training dynamics with F-GR as 
\begin{align}
\dot{w}_{t} &=- \nabla \mathcal{L}(w_t)  - \gamma \frac{\nabla \mathcal{L}(w_t+\varepsilon \nabla \mathcal{L}(w_t)) -\nabla \mathcal{L}(w_t) }{\varepsilon} \\
&= - q_1  \nabla \mathcal{L}(w_t) - q_2\nabla \mathcal{L} (w_t+\varepsilon \nabla \mathcal{L}(w_t)), \label{S2:0927}
\end{align}
where $q_1 = (1-\gamma/\varepsilon)$, $q_2 = \gamma/\varepsilon$. 
The training loss $ \mathcal{L}(w)$ is defined in (\ref{eq8:0927}).
The dynamics are rewritten as
\begin{align}
\frac{dw(t)}{dt}&= - \frac{q_1}{n} (\tilde{X}^{\top} r(t)) \circ w(t) - \frac{q_2}{n} (\tilde{X}^{\top} r^{*}(t)) \circ w^{*}(t), \label{S3:0928}
\end{align}
where $\circ$ denotes the element-wise product between vectors. 
We defined $r(t)=\tilde{X} w(t)^{2}-y$, $r^{*}(t)=\tilde{X} w^{*}(t)^{2}-y$, $w^*(t)=w(t)+\varepsilon \nabla \mathcal{L}(w(t))$, and put $\tilde{X}=\left[\begin{array}{ll}
X & -X
\end{array}\right] \in \mathbb{R}^{n \times 2 d}$. We recall that the square of the vector is an element-wise square operation. 
The general solution of (\ref{S3:0928}) is written as 
\begin{align}
    w(t) &= \begin{bmatrix} \alpha_0 \\  \alpha_0 \end{bmatrix} \circ \exp \left(- \frac{1}{n} \tilde{X}^{\top} \int_{0}^{t} (q_1 r(s) + q_2 r^*(s)) d s\right) \nonumber \\ 
    &\circ \exp \left(-\frac{q_2 \varepsilon}{n^2} \int_{0}^{t}(\tilde{X}^{\top} r^*(s)) \circ(\tilde{X}^{\top} r(s)) d s\right). \label{S6:0923}
\end{align}
This recovers the GD solution obtained by  \citet{woodworth2020kernel} for $(q_1,q_2)=(1,0)$, and SAM solution by \citet{andriushchenko2022towards} for $(q_1,q_2)=(0,1)$. GR requires us to consider general $(q_1,q_2)$.

 From Eq. (\ref{S6:0923}), we can represent an interpolation solution by 
\begin{align}
\beta_\infty&=w_{+}(\infty)^{2}-w_{-}(\infty)^{2} \\
&=2 \alpha^{2}_{F\text{-}GR} \circ \sinh \left( X^{\top} \nu \right), \label{S6:0126}
\end{align}
where $\nu=-\frac{2}{n} \int_{0}^{\infty} (q_1 r(s)+q_2 r^*(s)) d s$ and 
\begin{equation}
    \alpha_{GR}:=  \alpha_0 \circ  \exp \left(-\frac{\gamma}{n^2} \Psi \right), \ \ \Psi := \int_{0}^{\infty}\left({X}^{\top} r^*(s)\right) \circ\left({X}^{\top} r(s)\right) ds. \label{S8:0929}
\end{equation}
Put $ \beta_{\infty}=B_{\alpha_{GR}}\left(X^{\top} \nu\right)$ with $B_{\alpha_{GR}}(z)=2 \alpha^{2}_{F\text{-}GR} \circ \sinh (z)$. Because the form of the function $ \beta_{\infty}=B_{\alpha} \left(X^{\top} \nu\right)$ is 
the same as in the analysis of usual gradient descent \citep{woodworth2020kernel}, we can use exactly the same transformation of $\beta_\infty$ as it is. Their transformation is summarized as follows:  
suppose an interpolation solution $\beta_\infty$ 
written in the form of 
\begin{equation}
    \beta_{\infty}=\underset{\beta \in \mathbb{R}^{d} \text { s.t. } X \beta=y}{\arg \min } \phi(\beta).
\end{equation}
Then,  the KKT condition of the interpolation solution
is given by 
\begin{equation}
    \nabla_{\beta} \phi (\beta)=X^\top \nu,
\end{equation}
where $\nu$ is a Lagrange multiplier. 
Comparing this KKT condition and Eq. (\ref{S6:0126}), 
we can see that the function $\phi$ should satisfy
\begin{equation}
    \nabla_{\beta} \phi_{\alpha}(\beta)=B_{\alpha}^{-1}(\beta)=\operatorname{arcsinh}\left(\frac{1}{2 \alpha^{2}} \circ \beta\right).
\end{equation}
Taking the integral of $\nabla_{\beta} \phi_{\alpha}$, we obtain
\begin{equation}
    \beta_{\infty}({\alpha})=\underset{\beta \in \mathbb{R}^{d} \text { s.t. } X \beta=y}{\arg \min } \phi_{\alpha}(\beta)
\end{equation}
with
\begin{equation}
\phi_{\alpha}(\beta)=    \sum_{i=1}^{d} \alpha_{i}^{2} q\left(\beta_{i} / \alpha_{i}^{2}\right)
\end{equation}
and 
\begin{equation}
q(z)=2-\sqrt{4+z^{2}}+z \operatorname{arcsinh}(z / 2).
\end{equation}

While we have $\alpha=\alpha_0$ in the analysis of usual gradient descent \citep{woodworth2020kernel}, we 
have $\alpha=\alpha_{GR}$ for GR. Thus, the evaluation of GR reduces to that of $\alpha_{GR}$ and its exponent $\Psi$.

\subsubsection{Basic property of $\Psi$}

From the definitions of $r(t)$ and $r^*(t)$, we have
\begin{align}
    r^*(t)-r(t) &= \frac{2 \varepsilon}{n} \tilde{X}((\tilde{X}^{\top} r(t)) \circ w(t)^2)+\frac{\varepsilon^{2}}{n^2} \tilde{X}((\tilde{X}^{\top} r(t))^{2} \circ w(t)^{2}).
\end{align}
Then, 
\begin{align}
\Psi &= \int_{0}^{\infty}({X}^{\top} r(s))^2 d s + \frac{\varepsilon}{n} \int_{0}^{\infty}\underbrace{ 2({X}^\top\tilde{X}((\tilde{X}^{\top} r(s))\circ w(s)^2)  ) \circ({X}^{\top} r(s))}_{=:z(s)} d s \nonumber \\ 
&\quad + \frac{ \varepsilon^2}{n^2} \int_{0}^{\infty} \underbrace{ ({X}^\top \tilde{X}((\tilde{X}^{\top} r(s))^{2} \circ w(t)^{2})) \circ({X}^{\top} r(s))}_{=:{z}_h(s)} ds.
\end{align}
 Let us put
\begin{equation}
\Psi = \Psi_0 + \frac{\varepsilon}{n} \Psi_1 + \frac{ \varepsilon^2}{n^2} \Psi_2, \label{S17}
\end{equation}
\begin{equation}
\Psi_0 := \int_{0}^{\infty}({X}^{\top} r(s))^2 d s, \quad  \Psi_1 := \int_{0}^{\infty} z(s) d s, \quad
\Psi_2 := \int_{0}^{\infty} z_h(s) d s. \label{S18}
\end{equation}
Note that the first term $\Psi_0$ essentially corresponds to the implicit bias of the SAM update investigated in the previous study  \citep{andriushchenko2022towards}. Because the SAM update corresponds to $\gamma=\varepsilon$, the dominant term of $\gamma \Psi$ is $\Psi_0$ and $\Psi_1$ and $\Psi_2$ terms disappear. 
In our GR case, $\gamma$ and $\varepsilon$ have different scales in general and we need to evaluate these novel terms. The essential problem is that the positivity of $\Psi_1$ and $\Psi_2$ is  non-trivial. Fortunately, we can prove the positivity of $\Psi_1$ for a sufficiently small $\gamma$ in the following.

\subsubsection{Evaluation of exponents in $\alpha_{GR}$}
\label{secA13}

Theorem 4.3 is obtained from the following lemma.
\begin{lemma}
\label{lemmaA1}
Under Assumption 4.2 (i)-(iii), $\Psi_1 = nb(0)^2/2 +  \mathcal{O}(\gamma)$.
\end{lemma}

\noindent
{\it Proof of Lemma A.1.}
The dynamics (\ref{S3:0928}) are rewritten as 
\begin{align}
n\frac{dw}{dt} &= - \tilde{b}\circ w - \frac{\gamma}{n} [2 (\tilde{Z} (\tilde{b}\circ w^2) )\circ w + \tilde{b}^2 \circ w]  \nonumber \\
&\quad - \frac{\gamma \varepsilon}{n^2} [ (\tilde{Z}(\tilde{b}^2 \circ w^2))\circ w +  2(\tilde{Z}(\tilde{b}\circ w^2))\circ w \circ \tilde{b}]  -\frac{\gamma \varepsilon^2}{n^3} [(\tilde{Z}(\tilde{b}^2\circ w^2))\circ w \circ \tilde{b} ],
\end{align}
where we put $\tilde{b}=\tilde{X}^\top r$ and $\tilde{Z}=\tilde{X}^\top \tilde{X}$. This gives us
\begin{align}
 \frac{n}{2}\frac{d\beta}{dt} &= - {b}\circ a - \frac{\gamma}{n} \underbrace{[2 (Z({b}\circ a) )\circ a + {b}^2 \circ \beta]}_{=:Q_1(t)}  \nonumber \\
&\quad - \frac{\gamma \varepsilon}{n^2} \underbrace{[ ({Z}({b}^2 \circ \beta))\circ a +  2({Z}({b}\circ a))\circ \beta \circ {b}]}_{=:Q_2(t)} -\frac{\gamma \varepsilon^2}{n^3} \underbrace{[({Z}({b}^2\circ \beta))\circ \beta \circ {b} ]}_{=:Q_3(t)}, \label{eqS27:0907}
\end{align}
where we put $a=w_+^2+w_-^2$, ${b}={X}^\top r$ and $Z={X}^\top {X}$.
Note that $db/dt=X^\top (dr/dt) = X^\top X (d \beta/dt)$. 
By multiplying $X^\top X$ to (\ref{eqS27:0907}) and taking the Hadamard product with $b$, we have 
\begin{equation}
n  \frac{d b^2}{dt} = -  4 b \circ (X^\top X({b}\circ a))  -  \frac{4\gamma}{n} \underbrace{ b \circ[  X^\top X ( Q_1(t) +\frac{\varepsilon}{n}  Q_2(t) +\frac{ \varepsilon^2}{n^2}  Q_3(t) )]}_{=:Q(t)}.
\end{equation}
The point is that we have $2 b \circ (X^\top X({b}\circ a))=z(t)$. 
\com{This relation enables us to evaluate the seemingly complicated term $\Psi_1$ by the change of $b(t)^2$, which corresponds to a training loss.} 
By taking the integral over time, the above dynamics become
\begin{equation}
\Psi_1=\int_0^\infty z(s) ds = \frac{n}{2} b(0)^2 - 2 \frac{\gamma}{n} \int_0^\infty Q(s) ds. \label{S19:0124}
\end{equation}
We used  assumption (i) that we have a global minimum and $b(\infty)=0$.
If $\gamma$ is sufficiently small and $ \int_0^\infty Q(s) ds$ is finite, we will have a non-negative $\Psi_1$.

Here, we use  assumption (ii) that the parameter norm has a finite constant upper bound independent of $\gamma$ and $\varepsilon$. 
Because
$\|a(t)\|=\|w_+(t)^2+w_-(t)^2\| \leq \|w\|^2$, we have an upper bound of $\|a(t)\|$ as well:
\begin{equation}
 \|a(t)\| \leq \bar{a}.  \label{S36:0910}
\end{equation}
Define $\kappa_1 := \text{argmax}_i  \|X x^{(i)}\| $, $\kappa_2 := \text{argmax}_i \|x^{(i)} \|$ and $\kappa_3 :=  \|XX^\top \|_2$. Then, we find
\begin{align}
|Q_{1,i}(t)| &\leq 2a_i \|X x^{(i)}\| \|b \circ a\| + b_i^2 |\beta_i| \\
&\leq 2 \bar{a}^2 \kappa_1 \sqrt{\kappa_3} \|r(t)\|+\bar{a} \kappa_2^2 \|r(t)\|^2. \label{eqQ1}
\end{align}
where we used $\|b \circ a\|\leq \| b\|\|a\| \leq \sqrt{\kappa_3} \bar{a}\|r\|$ and  $\|\beta\| \leq \|a\| \leq \bar{a}$. Similarly, we have
\begin{align}
|Q_{2,i}(t)|  
&\leq  \bar{a}^2 \kappa_1 \kappa_3 \|r\|^2 +2 \bar{a}^2 \kappa_1 \kappa_2 \sqrt{\kappa_3} \|r\|^2, \label{eqQ2}
\end{align}
where we used $\|b^2\| \leq \sqrt{\sum_i (X_i r)^4} \leq {\sum_i (X_i r)^2} = \|b\|^2$. We also have
\begin{equation}
|Q_{3,i}(t)| \leq \bar{a}^2 \kappa_1 \kappa_2  \kappa_3 \|r\|^3.
\end{equation}
Note that under assumption (ii),
the training loss is upper-bounded as well because
\begin{align}
\|r(t)\| &\leq \|X\beta\|+\|y\| \leq \sqrt{\kappa_3} \bar{a} + \|y\| =: \bar{\mathcal{L}}.
\end{align}
Therefore, we have 
\begin{equation}
    |Q_{3,i}(t)| \leq \bar{a}^2 \kappa_1 \kappa_2  \kappa_3 \bar{\mathcal{L}} \|r\|^2. \label{eqQ3}
\end{equation}
After all, the inequalities (\ref{eqQ1},\ref{eqQ2},\ref{eqQ3}) lead to  
\begin{align}
\int_0^\infty ds Q_i(s)  &\leq    C \int_0^\infty ds  \|r\|^2  
\leq C \bar{R} \label{S34:0928}
\end{align}
where $C$ denotes an uninteresting positive constant, which depends on $\{\kappa_1, \kappa_2, \kappa_3, \bar{\mathcal{L}},\bar{a} \}$,  and we used the assumption (iii). After all, since the integral of $Q_i$ is bounded by a constant, we have  
\begin{equation}
\Psi_1= \frac{n}{2} b(0)^2 + \mathcal{O} (\gamma)
\end{equation}
for sufficiently small $\gamma$.
\QEDA

After all, putting 
$c_0 = \Psi_0/n^2$ and $c_2=\Psi_2/n^4$ and substituting $\Psi_1= \frac{n}{2} b(0)^2 + \mathcal{O} (\gamma)$  into Eq. (\ref{S8:0929}), we obtain Theorem 4.3.

\subsection{Derivation of Proposition 4.4}
\label{Sec_A3}

Consider the $i$-th entry satisfying $b_i(0) \neq 0$. This condition is rational because $b_i(0)$ is determined by the training error at initialization, that is, $X^\top \beta(0) -y$, and expected to take a positive value. 
First, from Ineq. (\ref{S34:0928}), we find
\begin{equation}
\Psi_{1,i} \geq \frac{3 n b_i(0)^2}{8} > 0 \ \ \text{for} \ \ \gamma \leq \frac{n^2b_i(0)^2}{16 C \bar{R}}. \label{S35:0928}
\end{equation}
Next, we evaluate $\Psi_2$.
Since  
\begin{equation}
z_h= (Z (b^2 \circ \beta))   \circ b, 
\end{equation}
we have
\begin{align}
|z_{h,i}| &\leq \kappa_1 \kappa_2 \kappa_3 \bar{a} \bar{\mathcal{L}} \|r\|^2.  
\end{align}
Therefore, 
\begin{equation}
|\Psi_{2,i}|=|\int_0^\infty z_{h,i}(s) ds| \leq C_h \bar{R},
\label{S32:0124}
\end{equation}
where $C_h$ denotes an uninteresting positive constant $4n\kappa_1 \kappa_2 \kappa_3 \bar{a} \bar{\mathcal{L}}$. 
Then, by using $\Psi_0 \geq 0$ and (\ref{S35:0928}), 
\begin{align}
\Psi_i &\geq  \frac{\varepsilon}{n} \Psi_{1,i} + \frac{ \varepsilon^2}{n^2} \Psi_{2,i} \geq  \varepsilon ( \frac{3b_i(0)^2}{8}  + \frac{ \varepsilon}{n^2} \Psi_{2,i}) \label{S35:0124}
\end{align}
for 
\begin{equation}
    \gamma \leq \min_i \frac{n^2 b_i(0)^2}{16C\bar{R}}  =: \gamma'. \label{gamma_dash}
\end{equation}
Furthermore, from (\ref{S32:0124}), we have 
\begin{align}
 \Psi_i &\geq  \varepsilon (\frac{3b_i(0)^2}{8}  - \frac{ \varepsilon}{n^2} C_h \bar{R})  \geq  \varepsilon  \frac{b_i(0)^2}{4}
\end{align}
for 
\begin{equation}
    \varepsilon \leq \min_i \frac{n^2 b_i(0)^2}{8C_h \bar{R}}  =: \varepsilon'. \label{var_dash}
\end{equation}
After all, we obtain $\alpha_{GR,i} \leq \alpha_{0,i} \exp (-\gamma \varepsilon c_{1,i}/2)$.
 \QEDA

As a side note, the inequality (\ref{S35:0928}) of $\gamma$  gives us some insight into non-asymptotic evaluation on how large $\gamma$ we can take. First, the constant $C$ includes $\bar{a}$ and it implies that we need a smaller $\gamma$ for a larger parameter norm $\bar{a}$. Second, note that $\bar{R}$ controls the integral of the training loss over the whole training dynamics. We need a smaller $\gamma$ as well for a larger $\bar{R}$ which implies the convergence of dynamics is slower. In the same way, we need a smaller $\varepsilon$ for larger $\bar{a}$ and $\bar{R}$.

\noindent
{\bf Remark on  an average of $\Psi_0$:} Note that we used no information of $\Psi_0$  in the inequality (\ref{S35:0124}). If one can make a tight lower bound of $\Psi_{0,i}$, it may improve the upper bound of $\alpha_{GR,i}$. Here, let us look at the average value of $\Psi_0$, that is, $\|\Psi_0\|_1 = \sum_{i=1}^d \Psi_{0,i}$. 
Suppose that $\int_0^\infty \mathcal{L}(w(t)) dt$  has a constant  lower bound $\underline{R}$. Then, we have
\begin{align}
\|\Psi_0\|_1 &=  \int_{0}^{\infty} r(s)^\top  ( XX^\top) r(s) ds \\ &\geq 4n \lambda_{min}(XX^\top) \underline{R}.
\end{align}
Although it seems not easy to obtain a lower bound of each $\Psi_{0,i}$, it is related to $\underline{R}$ on average.

\noindent
{\bf Case of B-GR (Derivation of Ineq. (\ref{eq16:0124})):} 
Note that we can see B-GR as the F-GR with a negative $\varepsilon$. For B-GR, instead of (\ref{S35:0124}), we have
\begin{equation}
\Psi_i \leq  \Psi_{0,i} + \varepsilon \frac{3b_i(0)^2}{8}  + \frac{ \varepsilon^2}{n^2} \Psi_{2,i}
\end{equation}
for $\gamma \leq \gamma'$.
By taking $-\varepsilon'\leq \varepsilon<0$, we have 
\begin{equation}
    \Psi_i \leq  \Psi_{0,i} +  \varepsilon b_i(0)^2/2.
\end{equation}
In addition, we have
\begin{align}
 \Psi_{0,i}  &\leq \kappa_2^2 \int_0^\infty \|r(s)\|^2 ds \leq 4n \kappa_2^2 \bar{R}.
\end{align}
By putting $D= \exp(- 4\kappa_2^2 \bar{R}/n)$, we obtain the result.

\subsection{Validity of Assumptions}
\label{SecA2}

Let us summarize the assumptions that we used in the above analysis.  
\begin{assumption}[Assumption 4.2 restated]
(i) the gradient dynamics converges to the interpolation solution satisfying $X\beta=y$, (ii) $\|w(t)\|$ has a constant upper \com{bound} independent of $\gamma$ and $\varepsilon$, (iii) for sufficiently small $\gamma$ and $\varepsilon$,  the integral of the training loss, i.e., \com{$\int_0^\infty \mathcal{L}(w(t)) dt$}, has a  constant upper bound $\overline{R}$ independent of $\gamma$ and $\varepsilon$.    
\end{assumption} 
These assumptions seem rational in the following sense. First, assumption (i) is commonly used in the study of DLNs \cite{woodworth2020kernel}. Second, \citet{nacson2022implicit} recently reported that
we can obtain interpolation solutions with a smaller parameter norm $\|w(t)\|$ using the discrete update with a larger learning rate. Because the interpolation solutions of gradient descent are also those of our learning with GR, assumption (ii) seems rational. 
The upper bound of assumption (iii) means that the convergence speed of $\mathcal{L}(w(t))$ does not get too small for sufficiently small $\gamma$ and $\varepsilon$. 
As a side note, we can replace assumption (iii) with the positive definiteness of a certain matrix. This is seemingly rather technical, but related to a sufficient condition that the dynamics converge to the global minima as follows.
\begin{assumption}[Alternative to Assumption 4.2 (iii)]
\label{As32}
For sufficiently small $\varepsilon$ and $\gamma$, the smallest eigenvalue of 
$S(t):=   X\text{diag}(a(t))X^\top$ is positive.
\end{assumption}
Since we suppose the overparameterized case ($d>n$), the matrix $X$ is a wide matrix and $S$ has no trivial zero eigenvalue. The positive definiteness of $S$ is a sufficient condition of global convergence as follows.
From Eq. (\ref{eqS27:0907}), we have
\begin{equation}
 \frac{n}{4}\frac{d\|r\|^2}{dt}  = \frac{n}{2} b^\top \frac{d\beta}{dt} = - r^\top S r   - \frac{\gamma}{n} r^\top X ( Q_1(t) +\frac{\varepsilon}{n}  Q_2(t) +\frac{ \varepsilon^2}{n^2}  Q_3(t) ). \label{S38:0929}
\end{equation}
Using the inequalities (\ref{eqQ1},\ref{eqQ2},\ref{eqQ3}), we have 
\begin{equation}
 \frac{n}{4} \frac{d\|r\|^2}{dt} \leq - \lambda_{min}^* \|r\|^2  + \gamma C \|r\|^2.
\end{equation}
where we take the lower bound of the smallest eigenvalue as 
$\lambda_{min}^*=\min_{t,\gamma,\varepsilon} \lambda_{min}(S(t)).$
By taking a sufficiently small $\gamma$ such that $\gamma < 3\lambda_{min}^*/(4C)$,  we obtain   
\begin{equation}
 \|r(t)\|^2 \leq  \|r(0)\|^2 \exp(-\lambda_{min}^* t/n),
\end{equation}
from Gr\"onwall's inequality.
Since $\mathcal{L}(w(t))= \|r(t)\|^2/(4n)$, we obtain global convergence. 
In addition, we have 
\begin{equation}
\int_0^\infty ds \|r(s)\|^2 \leq   \|r(0)\|^2 \int_0^\infty ds \exp(-\lambda_{min}^* t/n) = n  \|r(0)\|^2/\lambda_{min}^*.
\end{equation}
This gives the upper bound $\bar{R}$. 
Thus, instead of assumption (iii), we can apply Assumption C.3 in the transformation from (\ref{S34:0928}) to (\ref{S35:0928}).

Note that $S(t)$ is known as the neural tangent kernel in the lazy regime and its positive definiteness is straightforward \citep{woodworth2020kernel}.  
Although there is no proof of the positive definiteness in the rich regime, we observed it in numerical experiments and the assumption \ref{As32} seems rational.

\section{Derivation of Theorem 5.1}
\label{Sec_B}
It is straightforward to derive this theorem. 
Consider the time step $t$ satisfying  $\mathcal{L}(\theta_{t})<b$ and  $\mathcal{L}(\theta_{t+1})>b$. The update rule is given by 
\begin{align}
\theta_{t+1} &= \theta_{t} + \eta \nabla_\theta \mathcal{L}( \theta_{t}), \\
\theta_{t+2} &= \theta_{t+1} - \eta \nabla_\theta \mathcal{L}( \theta_{t+1}). 
\end{align}
Taking the summation, we get
\begin{align}
\theta_{t+2} &= \theta_t - \eta (\nabla_\theta \mathcal{L}( \theta_{t+1})-\nabla_\theta \mathcal{L}( \theta_{t}))\\
&= \theta_t -\eta^2 \frac{\nabla \mathcal{L}\left(\theta_t+\eta \nabla \mathcal{L}\left(\theta_t\right)\right)-\nabla \mathcal{L}\left(\theta_t\right)}{\eta}.
\end{align}
Similarly, for $\mathcal{L}(\theta_{t})>b$ and  $L(\theta_{t+1})<b$, we have
\begin{align}
\theta_{t+1} &= \theta_{t} - \eta \nabla_\theta \mathcal{L}( \theta_{t}), \\
\theta_{t+2} &= \theta_{t+1} + \eta \nabla_\theta \mathcal{L}( \theta_{t+1}). 
\end{align}
and get
\begin{align}
\theta_{t+2} &= \theta_t + \eta (\nabla_\theta \mathcal{L}( \theta_{t+1})-\nabla_\theta \mathcal{L}( \theta_{t}))\\
&= \theta_t -\eta^2 \frac{\nabla \mathcal{L}\left(\theta_t\right) -\nabla \mathcal{L}\left(\theta_t-\eta \nabla \mathcal{L}\left(\theta_t\right)\right) }{\eta}.
\end{align}

\end{document}